\documentclass[11pt,a4paper,titlepage,twoside]{memoir}
\setlrmarginsandblock{4cm}{3cm}{*}
\setulmarginsandblock{4cm}{3cm}{*}

\usepackage[OT1]{fontenc}
\usepackage[british]{babel}
\usepackage[utf8]{inputenc}
\usepackage[sc]{mathpazo}
\usepackage{amsmath,amssymb,amsfonts,mathrsfs}
\usepackage[amsmath,thmmarks]{ntheorem}
\usepackage{graphicx}
\usepackage{caption}
\usepackage{subcaption}
\usepackage{longtable}
\usepackage{wrapfig}
\usepackage{enumitem}
\usepackage{pdfpages}
\usepackage{float}
\usepackage{xfrac}
\usepackage[export]{adjustbox}
\usepackage{algorithm}
\usepackage[noend]{algpseudocode}
\usepackage[urldate=long,backend=biber,sorting=none,sortcites=true,natbib=true,style=numeric-comp]{biblatex}
\usepackage{multirow}

\usepackage{varioref}

\usepackage{datetime}

\usepackage{mathtools}

\usepackage[h]{esvect}

\usepackage{array}

\usepackage{pstricks,pst-all}

\usepackage{listings}
\usepackage{booktabs}

\nonzeroparskip
\defaultlists

\makeatletter

\if@twoside
  \copypagestyle{chapter}{Ruled}
\else
  \copypagestyle{chapter}{ruled}
\fi
\makeoddhead{chapter}{}{}{}
\makeevenhead{chapter}{}{}{}
\makeheadrule{chapter}{\textwidth}{0pt}
\copypagestyle{abstract}{empty}

\makechapterstyle{bianchimod}{%
  \chapterstyle{default}
  \renewcommand*{\chapnamefont}{\normalfont\Large\sffamily}
  
  \renewcommand*{\printchaptername}{%
    \chapnamefont\centering\@chapapp}

  }

\chapterstyle{bianchimod}

\setsecheadstyle{\Large\bfseries\sffamily}
\setsubsecheadstyle{\large\bfseries\sffamily}
\setsubsubsecheadstyle{\bfseries\sffamily}
\setparaheadstyle{\normalsize\bfseries\sffamily}
\setsubparaheadstyle{\normalsize\itshape\sffamily}
\setsubparaindent{0pt}

\captionnamefont{\sffamily\bfseries\footnotesize}
\captiontitlefont{\sffamily\footnotesize}
\setsecnumdepth{subsection}
\settocdepth{subsection}

\pretitle{\vspace{0pt plus 0.7fill}\begin{center}\HUGE\sffamily\bfseries}
\posttitle{\end{center}\par}
\preauthor{\par\begin{center}\let\and\\\Large\sffamily}
\postauthor{\end{center}}
\predate{\par\begin{center}\Large\sffamily}
\postdate{\end{center}}

\def\@advisors{}
\newcommand{\advisors}[1]{\def\@advisors{#1}}
\def\@department{}
\newcommand{\department}[1]{\def\@department{#1}}
\def\@thesistype{}
\newcommand{\thesistype}[1]{\def\@thesistype{#1}}

\renewcommand{\maketitlehookb}{\vspace{1in}%
  \par\begin{center}\Large\sffamily\@thesistype\end{center}}

\renewcommand{\maketitlehookd}{%
  \vfill\par
  \begin{flushright}
    \sffamily
    \@advisors\par
    \@department, UZH \& ETHZ
  \end{flushright}
}

\checkandfixthelayout

\makeatother

\theoremstyle{plain}
\numberwithin{equation}{chapter}

\theoremstyle{nonumberplain}
\theorembodyfont{\normalfont}
\theoremsymbol{\ensuremath{\square}}

\renewcommand{\epsilon}{\ensuremath\varepsilon}

\renewcommand{\phi}{\ensuremath{\varphi}}

\usepackage[linkcolor=black,colorlinks=true,citecolor=black,filecolor=black]{hyperref}
\usepackage{eso-pic}
\usepackage{xspace}
\usepackage[justification=centering]{caption}
\usepackage{siunitx}
\usepackage{tikz}
\usepackage{calc}
\usepackage{pgf}
\usepackage{fp}
\usetikzlibrary{shapes,arrows,positioning,shadows,calc,fit,arrows,decorations.pathreplacing,decorations.markings,decorations.pathmorphing,patterns,shapes,shapes.multipart}
\pgfdeclarelayer{bg}
\pgfsetlayers{bg,main}
\usepackage{listings} %For code in appendix

\definecolor{mygray}{rgb}{0.9,0.9,1.0}

\makeatletter
\def\BState{\State\hskip-\ALG@thistlm}
\makeatother

\usepackage{xparse}% http://ctan.org/pkg/xparse
\NewDocumentCommand{\ceil}{s O{} m}{%
	\IfBooleanTF{#1} % starred
	{\left\lceil#3\right\rceil} % \ceil*[..]{..}
	{#2\lceil#3#2\rceil} % \ceil[..]{..}
}

\makeatletter
\DeclareRobustCommand\onedot{\futurelet\@let@token\@onedot}
\def\@onedot{\ifx\@let@token.\else.\null\fi\xspace}

\makeatother

\usepackage{xcolor}	 % Required for custom colors
\definecolor{mygreen}{RGB}{44,85,17}
\definecolor{myblue}{RGB}{34,31,217}
\definecolor{mybrown}{RGB}{194,164,113}
\definecolor{myred}{RGB}{255,66,56}
\title{Tuning of Mixture-of-Experts Mixed-Precision Neural Networks}
\author{Fabian Tschopp}
\thesistype{Semester Project II}
\advisors{Supervisor: Dr.\ Matthew Cook}
\department{Institute of Neuroinformatics }
\date{June 11, 2018}

\bibliography{bib.bib}
\begin{document}

\frontmatter

\begin{titlingpage}
  \calccentering{\unitlength}
  \maketitle
\end{titlingpage}

%% The abstract of your thesis.  Edit the file as needed.
\begin{abstract}
Deep learning has become a useful data analysis method, however mainstream adaption in distributed computer software and embedded devices has been low so far. Often, adding deep learning inference in mainstream applications and devices requires new hardware with signal processors suited for convolutional neural networks.

This work adds new data types (quantized 16-bit and 8-bit integer, 16-bit floating point) to Caffe in order to save memory and increase inference speed on existing commodity graphics processors with OpenCL, common in everyday devices. Existing models can be executed effortlessly in mixed-precision mode. Additionally, we propose a variation of mixture-of-experts to increase inference speed on AlexNet for image classification.

We managed to decrease memory usage up to $3.29\times$ while increasing inference speed up to $3.01\times$ on certain devices.

We demonstrate with five simple examples how the presented techniques can easily be applied to different machine learning problems. 
The whole pipeline, consisting of models, example python scripts and modified Caffe library, is available as Open Source software.

\end{abstract}

\newpage
\renewcommand{\abstractname}{Acknowledgements}
\begin{abstract}

\subsection*{Family}
This work is dedicated to my father, Markus Tschopp (02.05.1958 - 07.11.2017). Without him, I would not have picked up programming back in 2004. Ever since he bought his first computer back in 1994, we spent a lot of time working on computers together.
	
\subsection*{University of Zurich, Institute of Neuroinformatics}
I would like to thank my supervisor Dr. Matthew Cook for his patience to await for the delayed project report.

\subsection*{Intel}
Intel generously sponsored an Iris Pro based Computer in 2016 to aid the development of OpenCL Caffe. Additionally, I would like to express my gratitude to Zhigang Gong for providing help on the 16-bit floating point implementation for Caffe.

\subsection*{AMD (Advanced Micro Devices)}
I would like to thank AMD and especially Greg Stoner for the generous hardware sponsoring of a Vega Founders Edition compute GPU in 2017, which empowered a lot of the development on the Caffe library and enabled me to use neural network models beyond what is possible on regular hardware.

\subsection*{Special Thanks}
My special thanks are extended to Dividiti, and especially Flavio Vella, Grigori Fursin and Anton Lokhmotov for their interest and exchange of valuable ideas on improving Caffe. 
Finally, I wish to thank Rahul Atlury for his enthusiastic interest and support of the project, and wish him success in his endeavor to write tutorials and a book on the use of OpenCL Caffe.

\end{abstract}

%% TOC with the proper setup, do not change.
\cleartorecto
\tableofcontents
\mainmatter
\chapter{Caffe Library Implementation}
\label{ch:caffelibrary}
\section{Introduction}
Caffe has originally been created by Yangqing Jia, Evan Shelhamer, and Jeff Donahue \cite{jia2014caffe}. Originally, Caffe was only intended for CPU and CUDA usage. We subsequently developed an OpenCL backend, based on ViennaCL \cite{RuppViennaCL}, to support a variety of commodity hardware in 2015 \cite{Tschopp2015,Tschopp2016,Caffe}.

Adaption for commodity hardware such as integrated GPUs, present in most modern computers, and embedded devices such as Raspberry Pi \cite{raspberrypi} and the Asus Tinkerboard \cite{asustinkerboard} has been low, however. This is in part due to too slow inference speeds, which is a task that would typically be carried out in end-user applications.

A possible usage scenario of our software would be to train a network on a discrete GPU for a robot, and then build the robot with a small, energy efficient embedded system-on-a-chip computer.

In this work, we attempt to increase inference speed on both desktop and mobile GPUs by adding lower precision (quantized 8/16-bit integer and 16-bit floating point) and mixed precision networks.

Additionally, we demonstrate how mixed-precision networks could potentially be combined with mixture-of-expert techniques to increase inference speed even further.

Important terminology used throughout this work:
\begin{itemize}
	\item \textbf{BLAS}: Basic linear algebra system: Matrix-matrix, matrix-vector, matrix-scalar, vector-vector and vector-scalar operations.
	\item \textbf{FP32}: 32-bit floating point (full or single precision).
	\item \textbf{FP16}: 16-bit floating point (half precision).
	\item \textbf{INT16}: Quantized 16-bit integer.
	\item \textbf{INT8}: Quantized 8-bit integer.
	\item \textbf{Caffe}: Refers to OpenCL Caffe implementation \cite{Caffe}.
	\item \textbf{LibDNN}: Our own cuDNN replacement library that provides a BLAS,\\ convolution- and pooling-operators for INT8, INT16, FP16 and FP32. It can be compiled and executed at runtime for all CUDA and OpenCL enabled devices.
	\item \textbf{cuDNN}: Convolution, pooling and auxiliary operator library for CUDA enabled GPUs \cite{CuDNN2014}.
	\item \textbf{CLBlast}: BLAS library for FP16 and FP32 for OpenCL enabled devices \cite{Nugteren2018}.
	\item \textbf{MOE}: Mixture-of-experts networks, where multiple sub-networks are added according to a gating-network \cite{Shazeer2017}.
	\item \textbf{GEMM}: General matrix-matrix multiplication.
\end{itemize} 

\section{Contributions}
The largest part of the work has been to improve the Caffe \cite{Caffe, BVLCCaffe} library to accommodate the lower precision floating point and quantized integer data types.

The changes can be grouped into adaptions on different levels, which amounted to the task of reprogramming Caffe almost from scratch, while keeping backward- and forward-compatibility with existing trained models and Python interface code.

The tasks in chronological order of implementation:
\begin{itemize}
	\item Merge existing efforts by Zhigang Gong on 16-bit floating point \cite{GongzgFP16}.
	\item Add CLBlast \cite{Nugteren2018} as drop-in OpenCL BLAS for 16-bit floating point.
	\item Remove the old dual-code path, where OpenCL and CUDA code was implemented separately in each Caffe layer.
	\item Add a new single-code path, where a generic C-based kernel can be written into strings inside each layer.
	These kernels will then be interpreted as either OpenCL or CUDA code and compiled at runtime (see Section \ref{subsec:device_abstraction}).
	\item Implement a caching system that stores precompiled kernels in a SQLite database so that subsequent execution of a network can load much faster. This was necessary to achieve the speed of the original CUDA implementation, where kernels are compiled at host code compile time. The new mode of compilation has several advantages (see Section \ref{subsec:device_abstraction}).
	\item Add new INT8 and INT16 data types.
	\item Include boilerplate code to use gemmlowp \cite{Jacob_2018_CVPR} for integer matrix multiplication. Gemmlowp has also been used to verify the LibDNN and Caffe native quantized operators for correctness (see Section \ref{sec:library_overview}).
	\item Implement quantization layers, quantization operators and quantized operators in Caffe and LibDNN (see Section \ref{sec:quantization}).
	\item Add the new data types to the Python interface to allow access from NumPy.
	\item Write Python interface examples for demonstrating and testing the mixture-of-experts mixed-precision neural networks (see Chapters \ref{ch:examples} and \ref{ch:benchmarks}).
	\item Implement a mixture-of-experts layer (see Section \ref{sec:mixture_of_experts}), which can run nested Caffe networks as gating- and expert-networks.
	\item Write usage examples for single-neuron (celsius-farenheit, Section \ref{sec:celsius_farenheit_example}), dual-fully-connected (MNIST, Section \ref{sec:mnist}), LeNet \cite{LeNet1998}, ImageNet \cite{AlexNet2012} and ImageNet-MOE (see Sections \ref{sec:mixture_of_experts} and \ref{ch:imagenet}) \cite{CaffeExamples}.
	\item Implement OpenMP support to preprocess the ILSVRC2012 images on the CPU faster, using multiple threads, before passing them to the network.
	\item Add a CMake cross-compile build system that allows compilation for ARM devices such as the Raspberry Pi 3 and the Asus Tinkerboard.
\end{itemize}

These changes sum up to 180'000 new and 170'000 removed lines of code over more than 150 commits of my own work, excluding merged contributions by others \cite{Caffe}.

\section{Caffe Software Architecture}
\label{sec:caffe_architecture}
\begin{figure}[H]
	\centering
	\includegraphics[width=0.8\textwidth]{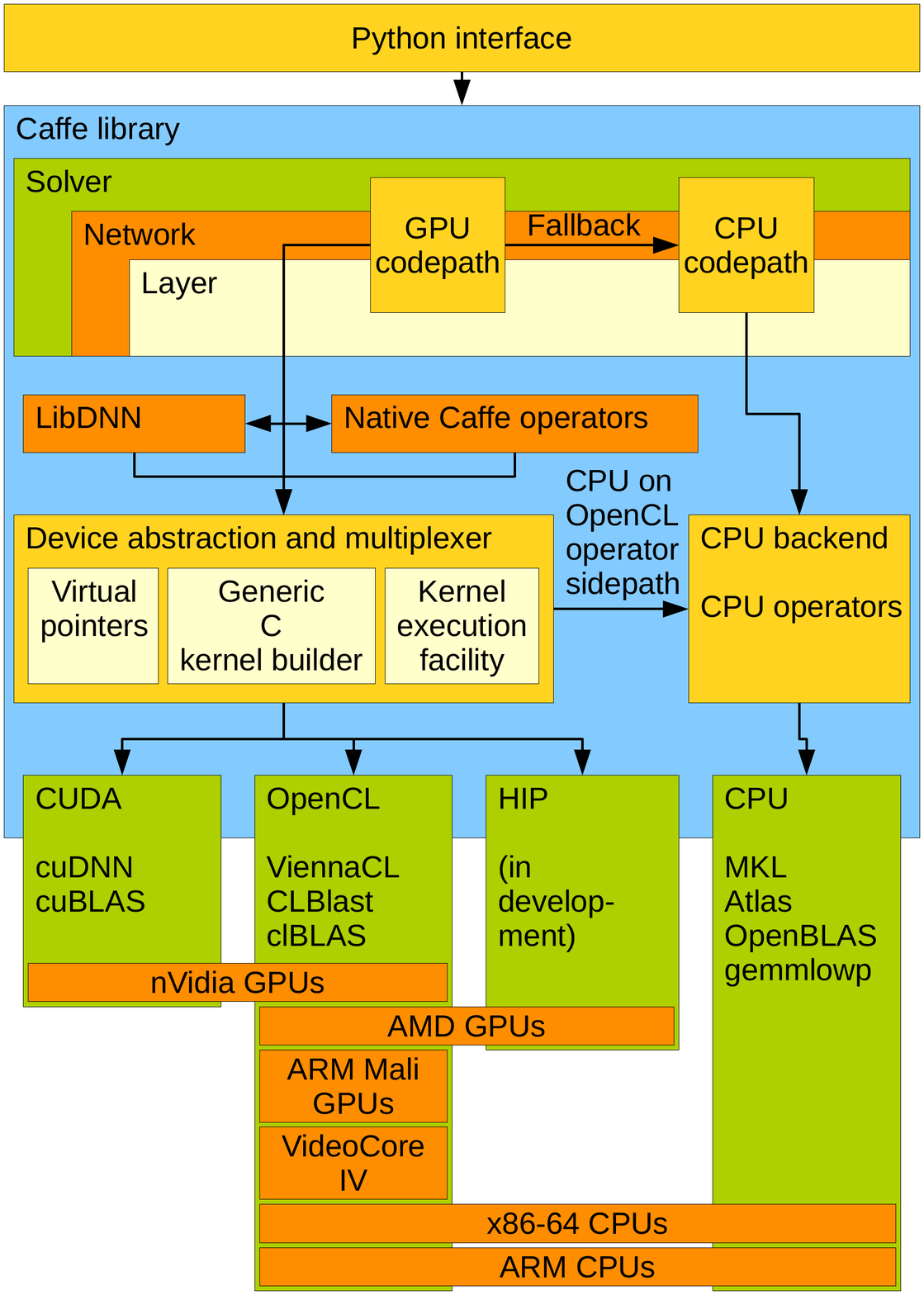}
	\caption{Caffe software architecture}
	\label{fig:caffe_architecture}
\end{figure}
Caffe is accessible through an easy to use Python interface (see Chapter \ref{ch:examples}). The Caffe library itself (written in C++), contains the three main elements users are interacting with: Solvers (also called optimizers), networks (graphs) and layers (operators). Layers store parameters, carry out computation and consume and produce blobs (tensors) in the process (see Figure \ref{fig:quantizer_layers}).

Computations within solvers, networks and layers can be executed on either the GPU or CPU backend. In this work, the GPU backend is unified into a single code path for both OpenCL and CUDA (see Section \ref{subsec:device_abstraction}). Many computations are offloaded to device specific external libraries, since these are often more optimized for the target device than the generic implementations inside Caffe (see Section \ref{sec:library_overview}).

\subsection{Library Overview}
\label{sec:library_overview}
Caffe depends on a number of external libraries to carry out common operators in a deep neural network. Not every library offers support for all operators (such as GEMM or convolution) or backends (CPU, OpenCL or CUDA), devices (ARM Mali, nVidia GeForce, AMD Polaris/Vega) or data types (FP16, FP32, INT8, INT16).
It is therefore essential to have a multiplexer within Caffe, which can decide on the most appropriate library for each case.

Generally, the Caffe multiplexer first looks for operators which are optimized for a certain device or backend. If no operator can be found, LibDNN operators can replace any convolution, pooling or BLAS operator for both CUDA and OpenCL (see Table \ref{tab:GPULibraries}).

\begin{table}[H]
	\begin{tabular}{p{0.19\textwidth}|p{0.1\textwidth}p{0.1\textwidth}p{0.1\textwidth}p{0.1\textwidth}p{0.1\textwidth}p{0.1\textwidth}}
		\hline
		Library & cuDNN & cuBLAS & CLBlast & ViennaCL & clBlas & LibDNN\\\hline\hline
		CUDA & yes & yes & no & yes, not used & no & yes\\
		OpenCL & no & no & yes & yes & yes & yes\\
		INT8 & no & no & no & no & no & yes\\
		INT16 & no & no & no & no & no & yes\\
		FP16 & partial, not used & partial, not used & yes & no & no & yes\\
		FP32 & yes & yes & yes & yes & yes & yes\\
		Convolution & yes & no & no & no & no & yes\\
		Pooling & yes & no & no & no & no & yes\\
		BLAS & no & yes & yes & yes & yes & yes\\
		Mobile GPUs & partial & partial & yes & partial & no & yes\\
		\hline
	\end{tabular}
	\caption{Features of GPU oriented libraries used in Caffe}
	\label{tab:GPULibraries}
\end{table}

Since there was no library offering quantized INT8 and INT16 types for all Caffe operators utilizing OpenCL yet, we had to program our own set of quantized operators within LibDNN. Convolutions can either be carried out in LibDNN, or through a transformation (im2col, col2im) plus a GEMM operator. For our benchmarks (see Chapter \ref{ch:benchmarks}), we always used convolutions and pooling through LibDNN, except where cuDNN and cuBLAS have faster alternatives and support the data type fully (see Table \ref{tab:GPULibraries}).

On CPUs, for which we do not assess performance in this work, a different set of libraries is used. Some CPUs also support OpenCL, however, the operators in OpenCL libraries are not optimized for CPUs at all. The Caffe multiplexer therefore chooses CPU libraries or Caffe native operators (see Table \ref{tab:CPULibraries}) even when run in OpenCL mode (see Figure \ref{fig:caffe_architecture}).

Most notably, with this work, we also included gemmlowp \cite{Jacob_2018_CVPR} for fast INT8 computations on various CPUs.

\begin{table}[H]
	\begin{tabular}{p{0.16\textwidth}|p{0.13\textwidth}p{0.13\textwidth}p{0.13\textwidth}p{0.13\textwidth}p{0.13\textwidth}}
		\hline
		Library & Atlas & OpenBLAS & MKL & gemmlowp & Caffe native\\\hline\hline
		INT8 & no & no & no & yes & yes\\
		INT16 & no & no & no & no & yes\\
		FP16 & no & no & no & no & yes\\
		FP32 & yes & yes & yes & no & yes\\
		Convolution & no & no & yes,\newline not used & no & yes\\
		Pooling & no & no & no & no & yes\\
		BLAS & yes & yes & yes & gemm only & yes\\
		\hline
	\end{tabular}
	\caption{Features of CPU oriented libraries used in Caffe}
	\label{tab:CPULibraries}
\end{table}

\subsection{Device Abstraction}
\label{subsec:device_abstraction}
Before this project, Caffe consisted of two GPU code paths: One for OpenCL and one for CUDA. This was no longer viable when adding new data types (FP16, INT16, INT8), since CUDA kernels are typically built into the host code and are also compiled with it. This requires the kernels to be fully configured before compile time (including C++ templates).
This would have meant to implement all four data types on two backends, resulting in up to eight code path variants plus additional CPU fallback code.

Because this was not a viable option to maintain, and due to advantages of generating layer parameter specific kernels at run-time, we decided to implement a fully device-abstracted backend to Caffe (see Figure \ref{fig:caffe_architecture}). This approach allows to have one code path for all data types and GPU backends.

We will explain the abstracted backend with the rectified linear unit (ReLU activation) layer as an example.

\subsubsection{Caffe Device Abstracted Host Code}

Each layer has three data types, which are declared as template (see Listing \ref{lst:host_launch_code} and \ref{lst:host_kernel_builder}, line 1):
\begin{itemize}
	\item \textbf{Dtype}: The compute data type. The compute type dictates the type of the trainable layer parameters and the internal computation precision.
	\item \textbf{MItype}: The bottom (input) data type to a layer. This dictates the data type of all bottom blobs consumed by a layer.
	\item \textbf{MOtype}: The top (output) data type to a layer. This dictates the data type of all top blobs generated by a layer.
\end{itemize}
It is left open to the layer implementation to choose which data type combinations are allowed as MItype, MOtype and Dtype. For ReLU, all types must be equal, but can be FP32, FP16, INT16 or INT8. Quantizer layers (see Section \ref{sec:quantization}) and MOE layers (see Section \ref{sec:mixture_of_experts}) can have a differing MItype and MOtype, but the Dtype (internal computation) is coupled to the MItype.

Additionally, the layer can derive two additional data types from the template types (see Listing \ref{lst:host_kernel_builder} and \ref{lst:host_launch_code}, lines 6-7):
\begin{itemize}
	\item \textbf{Difftype}: Difference type for quantized data types. This is typically a data type which is twice as large as the quantized type, and is signed. If such a large data type is not available, the largest supported signed integer type is used. While quantized data types are typically unsigned (see Section \ref{sec:quantization}), a Difftype must be able to store differences of the quantized type, which can become negative. For floating point data types, the Difftype is equivalent to the Dtype.
	\item \textbf{Acctype}: Accumulation type for quantized data types. This is a data type which accumulates the result of one or multiple Difftype or Dtype multiplication or addition results. Because the multiplication of two 16-bit integers can only fit into a 32-bit integer, the accumulation type is typically four times as large as the quantized type. This can seriously hinder the performance of quantized computation, as demonstrated in Chapter \ref{ch:benchmarks}.
\end{itemize}

\begin{lstlisting}[caption={Caffe device abstracted kernel builder code}, label={lst:host_kernel_builder}, language=C++]
template<typename Dtype, typename MItype, typename MOtype>
void ReLULayer<Dtype, MItype, MOtype>::GenerateProgram() {
	this->device_program_ = this->device_->CreateProgram();
	stringstream ss;
	
	typedef typename std::conditional<float_is_same<MItype>::value, MItype, typename std::conditional<sizeof(MItype) == 1, int16_t, typename std::conditional<sizeof(MItype) == 2, int32_t, int64_t>::type>::type>::type Difftype;
	typedef typename std::conditional<float_is_same<MItype>::value, MItype, typename std::conditional<sizeof(MItype) == 1, int32_t, int64_t>::type>::type Acctype;
	if (is_integer_type<MItype>()) {
		if (this->device_->template preferred_vector_width<int64_t>() > 0) {
			ss << this->device_program_->template define_vector_type<int64_t>(
			"Multtype", 0, 16);
		} else {
			ss << this->device_program_->template define_vector_type<int32_t>(
			"Multtype", 0, 16);
		}
	}
	
	ss << this->device_program_->setup();
	ss << this->device_program_->template define_type<Dtype>("Dtype");
	ss << this->device_program_->template define_type<MItype>("MItype");
	ss << this->device_program_->template define_type<MOtype>("MOtype");
	ss << this->device_program_->template define_type<Difftype>("Difftype");
	ss << this->device_program_->template define_type<Acctype>("Acctype");
	
	KernelArgs fw_args;
	fw_args.push_back(this->device_program_->template create_kernel_arg<uint_tp>("n", KERNEL_ARG_CONST));
	fw_args.push_back(this->device_program_->template create_kernel_arg<Dtype>("in", KERNEL_ARG_CONST | KERNEL_ARG_GLOBAL_MEM));
	fw_args.push_back(this->device_program_->template create_kernel_arg<Dtype>("out", KERNEL_ARG_GLOBAL_MEM));
	if (is_float_type<Dtype>()) {
		fw_args.push_back(this->device_program_->template create_kernel_arg<Dtype>("negative_slope", KERNEL_ARG_CONST));
	} else {
		fw_args.push_back(this->device_program_->template create_kernel_arg<int8_t>("shift_bits", KERNEL_ARG_CONST));
		fw_args.push_back(this->device_program_->template create_kernel_arg<Difftype>("in_zero", KERNEL_ARG_CONST));
		fw_args.push_back(this->device_program_->template create_kernel_arg<Acctype>("mult", KERNEL_ARG_CONST));
		fw_args.push_back(this->device_program_->template create_kernel_arg<int8_t>("shift", KERNEL_ARG_CONST));
		fw_args.push_back(this->device_program_->template create_kernel_arg<Acctype>("out_zero", KERNEL_ARG_CONST));
		fw_args.push_back(this->device_program_->template create_kernel_arg<Acctype>("out_min", KERNEL_ARG_CONST));
		fw_args.push_back(this->device_program_->template create_kernel_arg<Acctype>("out_max", KERNEL_ARG_CONST));
	}
	
	ss << this->device_program_->function("ReLUForward", fw_args);
	ss << this->device_program_->kernel_loop("uint_tp", "index", "n");
	if (is_float_type<Dtype>()) {
		ss << "out[index] = in[index] > (Dtype)0 ? in[index] : in[index]"
		<< " * negative_slope;"
		<< std::endl;
	} else {
		ss << "Difftype relu = max((Difftype)((Difftype)(in[index]) - "
		<< "in_zero), (Difftype)0);" << std::endl;
		ss << "Acctype reg = (Acctype)(((Multtype)(relu) * "
		<< "(Multtype)(mult)) / ((Multtype)1 << shift_bits));" << std::endl;
		ss << "if (shift >= 0) {" << std::endl;
		ss << "reg = reg >> shift;" << std::endl;
		ss << "} else {" << std::endl;
		ss << "reg = reg << -shift;" << std::endl;
		ss << "}" << std::endl;
		ss << "out[index] = (Dtype)(min(max(reg + out_zero, out_min), out_max));"
		<< std::endl;
	}
	ss << "}" << std::endl;
	ss << "}" << std::endl;
	this->device_program_->set_source(ss.str());
	this->device_program_->Compile(true, true);
}
\end{lstlisting}

The device abstracted kernel builder (Listing \ref{lst:host_kernel_builder}) allows to programmatically define the function arguments (lines 25-41).
The function arguments have several attributes that need to be defined:
\begin{itemize}
	\item A type, which will be adjusted to a valid OpenCL or CUDA type internally. The type can be a pointer or C data type. On OpenCL, FP16 arguments are only supported as pointers, but not values. Our device abstraction implementation will cast such values to FP32 on execution.
	\item A name, which is how the defined values can be addressed within the generated kernel. Pointer types on OpenCL are converted to a memory object plus an offset at which the data begins, since OpenCL does not have unified memory mapping. CUDA supports pointers directly, so the offset does not need to be passed to the kernel.
	\item Additional flags that declare if an argument is supposed to be a pointer to local memory, pointer to global memory or constant.
\end{itemize} 
The kernel name itself (in this case \textit{ReLUForward}) needs to be exclusive to the current compilation scope.

From lines 42 to 61, the actual kernel computation is defined. Since this code builds the kernel at runtime (lines 62-63), the required code for quantized integer computation (lines 48 to 58) or floating point (lines 44 to 46) can be selected at runtime, depending on the requested layer data types.

\begin{lstlisting}[caption={Caffe device abstracted host launch code}, label={lst:host_launch_code}, language=C++]
template<typename Dtype, typename MItype, typename MOtype>
void ReLULayer<Dtype, MItype, MOtype>::Forward_gpu(
				const vector<Blob<MItype>*>& bottom,
				const vector<Blob<MOtype>*>& top) {

	typedef typename std::conditional<float_is_same<MItype>::value, MItype, typename std::conditional<sizeof(MItype) == 1, int16_t, typename std::conditional<sizeof(MItype) == 2, int32_t, int64_t>::type>::type>::type Difftype;
	typedef typename std::conditional<float_is_same<MItype>::value, MItype, typename std::conditional<sizeof(MItype) == 1, int32_t, int64_t>::type>::type Acctype;
	
	vptr<const Dtype> bottom_data = bottom[0]->gpu_data();
	vptr<Dtype> top_data = top[0]->mutable_gpu_data();
	const int_tp count = bottom[0]->count();
	Dtype negative_slope = this->layer_param_.relu_param().negative_slope();
	
	shared_ptr<DeviceKernel> kernel = this->device_program_->GetKernel("ReLUForward");
	
	vector<size_t> work_size(1, count);
	vector<size_t> group;
	vector<size_t> local;
	
	this->device_->get_threads(&work_size, &group, &local, kernel.get(), true);
	
	kernel->add_arg(&count);
	kernel->add_arg(&bottom_data);
	kernel->add_arg(&top_data);
	if (is_float_type<Dtype>()) {
		kernel->add_arg(&negative_slope);
	} else {
		int8_t shift_bits =
		(this->device_->template preferred_vector_width<int64_t>() > 0 ? 32 : 16) / sizeof(MItype) - 1;
		Acctype mult;
		int8_t shift;
		QuantizerValues bottom_qv = this->bottom_quants_[0]->out_quantizer_values();
		QuantizerValues top_qv = this->top_quants_[0]->in_quantizer_values();
		QuantizerBase::ScaleQuantVals<Acctype>(&bottom_qv, &top_qv,
		&mult, &shift, shift_bits);
		Difftype bottom_zero = bottom_qv.get_zero<Difftype>();
		Acctype top_zero = top_qv.get_zero<Acctype>();
		Acctype top_min = top_qv.get_min<Acctype>();
		Acctype top_max = top_qv.get_max<Acctype>();
		kernel->add_arg(&shift_bits);
		kernel->add_arg(&bottom_zero);
		kernel->add_arg(&mult);
		kernel->add_arg(&shift);
		kernel->add_arg(&top_zero);
		kernel->add_arg(&top_min);
		kernel->add_arg(&top_max);
	}
	kernel->Execute(group, local);
}
\end{lstlisting}
In our kernel execution device abstraction (Listing \ref{lst:host_launch_code}), on the host code, the GPU memory is referenced with virtual pointers (\textit{vptr}, lines 9-10). These virtual pointers hide either a CUDA pointer or an OpenCL memory object plus offset.

When executing the generated and compiled kernels, the order of operations is as follows:
\begin{enumerate}
	\item The kernel is retrieved from the current compilation scope (line 14).
	\item Appropriate local and global work sizes (threads and thread groups) are selected according to the number of work items in the program (lines 16 - 20). This query is device and implementation dependent.
	\item The kernel arguments are populated (Listing \ref{lst:host_launch_code}, lines 22-46) in accordance to the function definition (Listing \ref{lst:host_kernel_builder}, lines 25-38).
	\item Computation is scheduled for execution on the current backend (line 48). The computation can run asynchronously to the host code.
\end{enumerate}

\subsubsection{Emitted OpenCL Runtime Code}

\begin{lstlisting}[caption={Emitted OpenCL code for FP32 ReLU}, label={lst:opencl_fp32_relu}, language=C++]
__kernel
void ReLUForward(const uint32_t n, __global const float* in, __global float* out, const float negative_slope) {
	for (uint_tp index = get_global_id(0); index < (n); index += get_global_size(0)) {
		out[index] = in[index] > (Dtype)0 ? in[index] : in[index] * negative_slope;
	}
}
\end{lstlisting}

\newpage
\begin{lstlisting}[caption={Emitted OpenCL code for INT8 ReLU}, label={lst:opencl_int8_relu}, language=C++]
__kernel
void ReLUForward(const uint32_t n, __global const uint8_t* in, __global uint8_t* out, const int8_t shift_bits, const int16_t in_zero, const int32_t mult, const int8_t shift, const int32_t out_zero, const int32_t out_min, const int32_t out_max) {
	for (uint_tp index = get_global_id(0); index < (n); index += get_global_size(0)) {
		Difftype relu = max((Difftype)((Difftype)(in[index]) - in_zero), (Difftype)0);
		Acctype reg = (Acctype)(((Multtype)(relu) * (Multtype)(mult)) / ((Multtype)1 << shift_bits));
		if (shift >= 0) {
			reg = reg >> shift;
		} else {
			reg = reg << -shift;
		}
		out[index] = (Dtype)(min(max(reg + out_zero, out_min), out_max));
	}
}
\end{lstlisting}

\subsubsection{Emitted CUDA Runtime Code}

\begin{lstlisting}[caption={Emitted CUDA code for FP32 ReLU}, label={lst:cuda_fp32_relu}, language=C++]
extern "C" __global__ void 
ReLUForward(const uint32_t n, const float* in, float* out, const float negative_slope) {
	for (uint_tp index = blockIdx.x * blockDim.x + threadIdx.x; index < (n); index += blockDim.x * gridDim.x) {
		out[index] = in[index] > (Dtype)0 ? in[index] : in[index] * negative_slope;
	}
}
\end{lstlisting}

\begin{lstlisting}[caption={Emitted CUDA code for INT8 ReLU}, label={lst:cuda_int8_relu}, language=C++]
extern "C" __global__ void 
ReLUForward(const uint32_t n, const uint8_t* in, uint8_t* out, const int8_t shift_bits, const int16_t in_zero, const int32_t mult, const int8_t shift, const int32_t out_zero, const int32_t out_min, const int32_t out_max) {
	for (uint_tp index = blockIdx.x * blockDim.x + threadIdx.x; index < (n); index += blockDim.x * gridDim.x) {
		Difftype relu = max((Difftype)((Difftype)(in[index]) - in_zero), (Difftype)0);
		Acctype reg = (Acctype)(((Multtype)(relu) * (Multtype)(mult)) / ((Multtype)1 << shift_bits));
		if (shift >= 0) {
			reg = reg >> shift;
		} else {
			reg = reg << -shift;
		}
		out[index] = (Dtype)(min(max(reg + out_zero, out_min), out_max));
	}
}
\end{lstlisting}

When looking at the emitted code (Listings \ref{lst:opencl_fp32_relu}, \ref{lst:opencl_int8_relu}, \ref{lst:cuda_fp32_relu}, \ref{lst:cuda_int8_relu}), it is apparent that the computation part is only different between the FP32 and INT8 versions, where FP32 additionally supports a negative slope, while INT8 requires to offset and shift the operands to do the quantized operation correctly (see Section \ref{sec:quantization}).

CUDA and OpenCL, since both are based on the C language, do not differ significantly, which is why it makes sense to have a device abstracted code generator.
Semantically, OpenCL and CUDA declare the function slightly different, with additional qualifiers before function and argument names. This is handled by the kernel builder, within the same macro that defines the kernel function, given a name and a list of arguments (Listing \ref{lst:host_kernel_builder}, line 41).

The code generator also automatically selects the proper way to retrieve the concurrent thread indices (Listing \ref{lst:host_kernel_builder}, line 42), which is different for CUDA and OpenCL (see Listing \ref{lst:opencl_fp32_relu}, line 3 and Listing \ref{lst:cuda_fp32_relu}, line 3).

Conclusively, the device abstracted backend allows the Caffe code base to remain small, hierarchical and easy to maintain. Code duplication is kept minimal, and all data types and devices are serviced from a single, unified host code. Adding additional backends such as HIP \cite{amdhip} in the future will be easy, and will not require modification to most of the Caffe layers, networks or solvers (see Figure \ref{fig:caffe_architecture}).
\section{Quantization}
\label{sec:quantization}
Quantized neural networks have gained increasing popularity due to their reduced memory and computation footprint (see Section \ref{sec:memory_consumption}). Often, INT8 and INT16 or even lower precisions still deliver close to the same accuracy as networks using FP32 or PF16 data types (see Section \ref{sec:inference_accuracy}).

Our implementation of INT8 closely follows gemmlowp \cite{Jacob_2018_CVPR}. However, gemmlowp is only suitable for CPU execution and optimized heavily for instruction sets found within common ARM and x86-64 processors. The library cannot be used for either CUDA or OpenCL.

We currently only support a quantized forward (inference) pass in Caffe. Training at lower precisions is not implemented. If a model trained in FP32 needs quantization, the necessary parameters can be estimated (see Section \ref{subsec:parameter_estimation}) and allow lower precision execution thereafter. Additionally, quantizer operators support pseudo-quantization, where the network computes at FP32 precision, but every blob (tensor), that is scheduled for quantization, is binned to the appropriate amount of distinct values (255 values for INT8 or 65535 values for INT16).

\subsection{Quantization Integration to Caffe}
\label{subsec:quantization_caffe}
Since we wanted to make quantization in Caffe intuitive for the users (see Section \ref{ch:examples}), we had to find a way to retrofit it into Caffe, which makes all network models forward- and backward-compatible with existing trained networks.

\begin{figure}[H]
	\centering
	\includegraphics[width=0.7\textwidth]{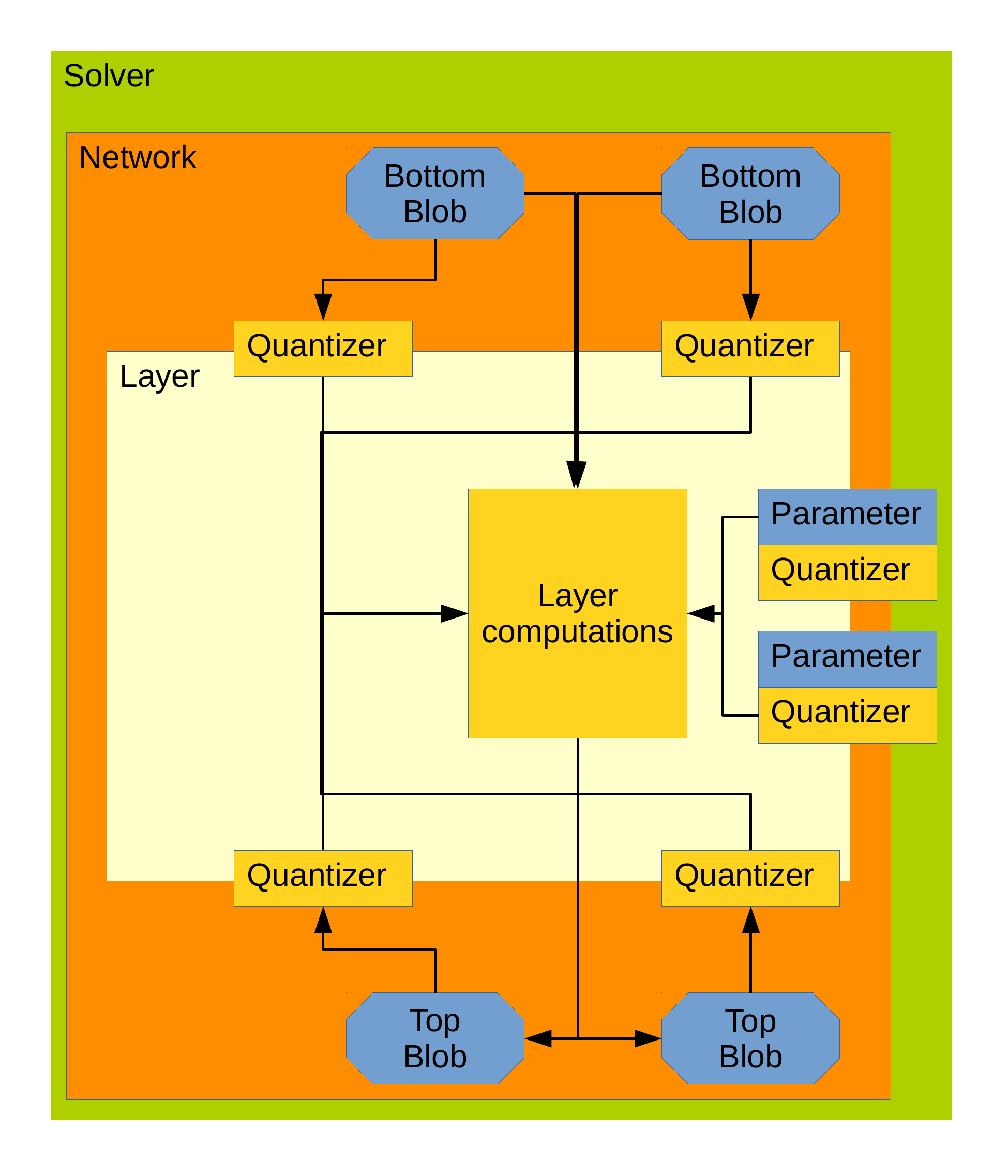}
	\caption{Caffe layer architecture with newly introduced quantizers.}
	\label{fig:quantizer_layers}
\end{figure}

The solution we chose was to add quantizer objects at strategic points within Caffe. Every layer now has a set of quantizers integrated (see Figure \ref{fig:quantizer_layers}):
\begin{itemize}
	\item Every bottom (input) blob to a layer has a quantizer within the layer associated with it. It mediates between the MItype and Dtype of a layer.
	\item Every top (output) blob to a layer has a quantizer within the layer associated with it. It mediates between the Dtype and MOtype of a layer.
	\item Every trainable set of network parameters has a quantizer within the layer associated with it. It mediates between the Dtype of a layer and the Dtype of the solver and network.
\end{itemize}

A quantizer object has several tasks, depending on the location of the quantizer, current precision and state of the layer:
\begin{itemize}
	\item During training in FP32 precision, the quantizers are either passive or perform pseudo-quantization on the bottom and top blobs.
	\item When testing in FP32 mode, the quantizers are either passive or in observational mode. When observing, the quantizer records the maximum and minimum observed values throughout inference, which are then used to estimate quantization parameters for a subsequent inference at lower precisions (see Section \ref{subsec:parameter_estimation} and Chapter \ref{ch:examples} for usage examples).
	\item During reduced-precision inference (INT8, INT16), all quantizers provide the layer computations with quantization parameters such as value range, scale, and offsets (see Section \ref{subsec:quantized_operations}). An example of this particular use case can be found in Listing \ref{lst:host_launch_code}.
	\item Optionally, the layer computations are allowed to use the quantizers to convert data from MItype to Dtype, Dtype to MOtype and vice-versa. This feature is currently only utilized by the quantization layer, which sits between two layers with differing input- and output types and connects them by converting the blobs to the appropriate type.
\end{itemize}

\subsection{Parameter Estimation}
\label{subsec:parameter_estimation}
Quantization parameters need to be estimated from the maximum ($f_{max}$) and minimum ($f_{min}$) of the observed floating point values per blob (tensor). Additionally, zero ($i_0$) and one ($i_1$) should be representable as accurately as possible, because otherwise we might introduce an unwanted additive and multiplicative bias \cite{Jacob_2018_CVPR}.

First, the scale $s$ is estimated, combining the range of floating point values observed ($f_{max}-f_{min}$) with the representable range of the quantized type ($i_{max}-i_{min}$). For INT8, we have $i_{max}=255$ and $i_{min}=0$:
\begin{equation}
	s = \frac{f_{max}-f_{min}}{i_{max}-i_{min}}
\end{equation}

With the scale, we can then estimate the zero-point $i_0$:
\begin{equation}
	i_0 = \min(\max(\lfloor i_{min} - \frac{f_{min}}{s}\rceil,i_{min}),i_{max})
\end{equation}

And the one-point  $i_1$ is trivially defined as:
\begin{equation}
	i_1 = \frac{1}{s} + i_0
\end{equation}

The parameters are estimated in full precision and are cast to the quantized data type before being passed to quantized operators within Caffe.

\subsection{Quantized Operations}
\label{subsec:quantized_operations}
Quantized operations sometimes require additional operations. The most usual example is multiplication, because multiplication of values with different quantization parameters requires adjusting the offset and scale of the result, which can then be stored in yet again another format with different parameters. Our implementation is equivalent to gemmlowp \cite{Jacob_2018_CVPR}, but rearranges the order of operators slightly to suit GPUs rather than CPUs.

We define a simple multiplication in full precision as:

\begin{equation}
	c = a \cdot b
\end{equation}

If we want to carry out the equivalent multiplication in quantized types, the operation becomes:

\begin{equation}
	c = ((a - i^{(a)}_0) \cdot (b - i^{(b)}_0)) \cdot \frac{s^{(a)}\cdot s^{(b)}}{s^{(c)}} + i^{(c)}_0
\end{equation}

where $i^{(a)}_0$, $i^{(b)}_0$ and $i^{(c)}_0$ are the respective estimated zero-points of $a$, $b$ and $c$. The scale terms $s^{(a)}$, $s^{(b)}$, $s^{(c)}$ rescale the result into the target domain.

The computation can be separated, which is done in our GEMM and convolution kernels, as well as in gemmlowp \cite{Jacob_2018_CVPR}, and allows to use faster code paths such as nVidia DP4A/DP2A \cite{cudadp4dp2}. The faster code can be used because, as we have seen in Section \ref{subsec:device_abstraction}, differences for an INT8 operation are stored in INT16, while sums of multiplications are stored in INT32. If we compute differences before multiplication, then the multiplication has to be carried out in INT16. But when executing the multiplications first, they can be done in INT8 and accumulated into either INT16 or INT32, depending on the expected size of the result sum.

Changing the order of operations also reduces the number of computations in matrix multiplications trivially, since all $i_0$ remain constant during the operation, and only $a$, $b$ and $c$ change throughout the rows and columns of a matrix.

\begin{equation}
	c = (a \cdot b - b \cdot i^{(a)}_0 - a \cdot i^{(b)}_0 + i^{(a)}_0 \cdot i^{(b)}_0) \cdot \frac{s^{(a)}\cdot s^{(b)}}{s^{(c)}} + i^{(c)}_0 
\end{equation}

Unfortunately, since we want to carry out all operators as integer types only, rescaling with $\frac{s^{(a)}\cdot s^{(b)}}{s^{(c)}}$ is not an option. Instead, we search an integer multiplier $m$ and integer arithmetic right shift value $h$ that fulfill the following equation:
\begin{equation}
	(x \cdot m) \gg h =  x * \frac{s^{(a)}\cdot s^{(b)}}{s^{(c)}}
\end{equation}

The shift and multiplier values can be determined once and remain constant. The equation can be fulfilled because an arithmetic shift is equivalent to dividing by a power of two. The resulting operation becomes:
\begin{equation}
	c = ((a \cdot b - b \cdot i^{(a)}_0 - a \cdot i^{(b)}_0 + i^{(a)}_0 \cdot i^{(b)}_0) \cdot m) \gg h  + i^{(c)}_0 
\end{equation}

In practice, additional rounding and clipping operations are in place to ensure no overflow or bias are introduced into the neural network. We omit detailed explanation of these additional steps, because they are use-case dependent and vary greatly from operator to operator. These operations can be found in the Caffe operators in the source code \cite{Caffe}.

\section{Mixture of Experts}
\label{sec:mixture_of_experts}
Mixture of experts neural networks combine different sub-networks in a larger network. These have typically been used to save inference time, add more parameters to a model or make multi-GPU training faster and easier \cite{Shazeer2017}. MOE networks can be seen as member of a class of networks that have divergent computation paths. Other examples of such networks include networks with conditional computations and early exits \cite{Bengio2015}.

For our work, we chose to use MOE networks to accelerate inference on low-end devices such as the Intel Iris Pro 540 and ARM Mali T764 (see Section \ref{sec:devices}). These devices profit from selectively processing more, but smaller neural network operators.

The mixture-of-experts network consists of four parts:
\begin{itemize}
	\item The main network, which is a regular convolutional neural network before and after the mixture-of-experts stage.
	\item A collection of experts with identical input and output tensor dimensions.
	\item A gating network, of which we take the output to decide which experts to use, usually combined with a noise and softmax stage.
	\item A gating selection stage, which selects the most useful experts and mixes their output. It is important to note that the experts are selected differently for every element of a batch during minibatch training and inference, since the choice of expert depends on a single sample only.
\end{itemize}

Unfortunately, the gating selection stage reinforces good experts to be used more often, which means after some training time, a few experts are very well trained, while others are untrained and completely disabled. A similar problem has been noted in previous work \cite{Shazeer2017}.

To mitigate the problem, we use a noisy gating function, using additive and multiplicative noise, to compute probabilities $p_i$ for each expert:
\begin{equation}
	q_i = \exp({w^{(a)}_ix + w^{(b)}_ix\cdot\mathcal{N}(\mu=0, \sigma=1) + w_i^{(c)}\cdot\mathcal{N}(\mu=0, \sigma=10)})
\end{equation}
\begin{equation}
	p_i = \frac{q_i}{\sum_{j=1}^{N}q_j} \qquad \forall i \in [1, N]
\end{equation}
Where $x$ is the gating network output. The network is expected to start with noisy gating and should slowly select experts smarter through the trainable weights $w^{(a)}$, while decreasing the noise regulating weights $w^{(b)}$ and $w^{(c)}$ towards zero.

The probabilities $p_i$ are normalized so that they are non-zero only for selected experts and add up to one.
The network then only computes the top $K$ selected experts from $N$ total experts, and sums their output weighted by the normalized probabilities $p_i$.

Additionally, we use a $L2$ regularization loss on the discrete number of times an expert has been used, compared to the expected average if every expert is used equally often:
\begin{equation}
	l_2 = \frac{1}{N} \sum_{i=1}^{N} (\frac{K}{N}-\frac{c_i}{B})^2
\end{equation}
Where $c_i$ is the observed number of times expert $i$ has been used in the current batch of batch-size $B$.

Our ImageNet-MOE configuration is explained in Section \ref{sec:imagenet_moe_config}, with complete network graphs in Appendix \ref{sec:app_imagenet_moe}.
\chapter{Examples}
\label{ch:examples}
These examples demonstrate the ease of use of the new data types through the python interface.
Larger examples (LeNet and ImageNet) have been omitted in text-form, but can be found online \cite{CaffeExamples}.

\section{Celsius-Farenheit}
\label{sec:celsius_farenheit_example}
This example uses a single neuron to compute the farenheit value corresponding to a celsius value.
A single neuron is sufficient since this operation is linear:
\begin{equation}
	y = ax+b
\end{equation}
The network has to train in full precision (FP32) to arrive at the estimated parameters of $a=1.8$ and $b=32.0$.

\begin{lstlisting}[caption={Creating the celsius-farenheit network}, label={lst:celsius_farenheit_create}, language=Python]
import sys
sys.path.append('..')
from caffe_examples_setup import *

# We create the network for float, half, int16 and int8
data_types = [caffe.data_type.CAFFE_HALF, caffe.data_type.CAFFE_FLOAT,
caffe.data_type.CAFFE_INT8_QUANTIZED, caffe.data_type.CAFFE_INT16_QUANTIZED]
data_types_names = ['half', 'float', 'int8', 'int16']

for data_type in zip(data_types, data_types_names):
	net = caffe.NetSpec()
	net.celsius = L.Input(input_param=dict(shape=dict(dim=[1,1,1,1])), ntop=1)
	net.farenheit = L.Input(input_param=dict(shape=dict(dim=[1,1,1,1])), ntop=1, include=dict(phase=0))
	
	net.neuron = L.InnerProduct(net.celsius,
								bottom_data_type = data_type[0],
								compute_data_type = data_type[0],
								top_data_type = data_type[0],
								inner_product_param = dict(num_output = 1,
								weight_filler = dict(type='constant'),
								bias_filler = dict(type='constant')))
	net.output = L.Quantizer(net.neuron,
							 bottom_data_type = data_type[0],
							 compute_data_type = data_type[0],
							 top_data_type = caffe.data_type.CAFFE_FLOAT)
	net.euclidean = L.EuclideanLoss(net.output, net.farenheit, include=dict(phase=0))
	
	
	protonet = net.to_proto()
	protonet.name = 'net'
	with open(protonet.name + '_' + data_type[1] + '.prototxt', 'w') as f:
		print(protonet, file=f)
\end{lstlisting}
In the past, it was necessary to write each Caffe network as a protocol text. In more recent versions, programmatically creating networks has been made easy. In Listing \ref{lst:celsius_farenheit_create}, we create a single-neuron network for FP32, FP16, INT16 and INT8 data types and store them on disk. As described in Section \ref{subsec:quantization_caffe}, each layer now has additional parameters to describe its input (bottom), compute and ouput (top) data type. Our neuron is always using the selected precision (lines 16-18). The network output is always a floating point number, converted by a quantizer layer (lines 22-25). Network inputs and the loss are also computed at full precision.

\begin{lstlisting}[caption={Training the celsius-farenheit network}, label={lst:celsius_farenheit_train}, language=Python]
import sys
sys.path.append('..')
from caffe_examples_setup import *

# Choose the precision (half, float, int8 or int16).
precision = 'float'

# Define the training and testing data
values_celsius = np.array([(float)(c) for c in range(-273,1000)])
# We know that farenheit = celsius * 1.8 + 32.0
values_farenheit = np.array([c*1.8+32.0 for c in values_celsius])

# Split data into training (90%) and testing (10%)
indices = np.random.permutation(values_celsius.shape[0])
training_idx, test_idx = indices[:(int)(90*values_celsius.shape[0]/100)], indices[(int)(90*values_celsius.shape[0]/100):]

values_celsius_train = values_celsius[training_idx]
values_farenheit_train = values_farenheit[training_idx]
values_celsius_test = values_celsius[test_idx]
values_farenheit_test = values_farenheit[test_idx]

# Create a solver with a few typical parameters
# The solver will perform SGD on our data
solver_config = caffe.SolverParameter()
solver_config.train_net = 'net_' + precision + '.prototxt'
solver_config.base_lr = 1.0
solver_config.momentum = 0.99
solver_config.weight_decay = 0.00005
solver_config.lr_policy = 'inv'
solver_config.gamma = 0.01
solver_config.power = 0.75
solver_config.max_iter = 2000
solver_config.snapshot = 500
solver_config.snapshot_prefix = 'net'
solver_config.type = 'Adam'
solver_config.display = 1

# Do the training
losses = []

plt.ion()
plot_obj, = plt.plot(losses)
ax = plt.gca()
plt.show()
plt.pause(0.001)

solver = caffe.get_solver(solver_config)
for i in range(0, solver_config.max_iter):
	# Pick a random sample for training
	k = random.randint(0,len(values_celsius_train)-1)
	# Load the sample into the network
	solver.net.blobs['celsius'].data[0] = values_celsius_train[k]
	solver.net.blobs['farenheit'].data[0] = values_farenheit_train[k]
	# Train one step
	loss = solver.step(1)
	# Display the learning progress every 20 steps
	if (i % 100 == 0):
		losses.append(loss)
		plot_obj.set_data(range(0,len(losses)), losses)
		ax.relim()
		ax.autoscale_view(True, True, True)
		plt.draw()
		plt.pause(0.001)


# Run a few test steps to observe the value ranges (for quantization)
error = []
testnet = caffe.Net(str('net_' + precision + '.prototxt'), caffe.TEST, weights='net_iter_'+str(solver_config.max_iter)+'.caffemodel')
# Enable quantizer observation
testnet.quant_mode = caffe.quantizer_mode.CAFFE_QUANT_OBSERVE
for c,f in zip(values_celsius_test,values_farenheit_test):
	testnet.blobs['celsius'].data[0] = c
	testnet.forward()
	predicted_f = testnet.blobs['output'].data[0,0]
	print('Cesius: '+str(c)+'C, predicted: '+str(predicted_f)+' F, actual: '+str(f)+' F')
	error.append(abs(f-predicted_f))

print('Average error: '+str(np.array(error).mean())+' F')

# Store the network parameters, including obtain quantizer information
testnet.save('net_trained.caffemodel')
print("Done.")
\end{lstlisting}

In Listing \ref{lst:celsius_farenheit_train}, lines 1-64 describe the usual training process. Additionally, since we want to run the network in quantized mode later, we need to add lines 66-82. This part of the code runs a few test examples through the network at full precision and collects statistics about the value domain of each blob and parameter in the network (see Section \ref{sec:quantization}). The resulting values are stored as quantization parameters together with the trained network (line 81).

\begin{lstlisting}[caption={Testing the celsius-farenheit network}, label={lst:celsius_farenheit_test}, language=Python]
import sys
sys.path.append('..')
from caffe_examples_setup import *

# Choose the precision (half, float, int16 or int8)
precision = 'float'

# Define the training and testing data
values_celsius = np.array([(float)(c) for c in range(-273,1000)])
# We know that farenheit = celsius * 1.8 + 32.0
values_farenheit = np.array([c*1.8+32.0 for c in values_celsius])

# Split data into training (90%) and testing (10%)
indices = np.random.permutation(values_celsius.shape[0])
training_idx, test_idx = indices[:(int)(90*values_celsius.shape[0]/100)], indices[(int)(90*values_celsius.shape[0]/100):]

values_celsius_train = values_celsius[training_idx]
values_farenheit_train = values_farenheit[training_idx]
values_celsius_test = values_celsius[test_idx]
values_farenheit_test = values_farenheit[test_idx]

# Test how accurate the network has learned it's task
error = []
testnet = caffe.Net(str('net_' + precision + '.prototxt'), caffe.TEST, weights='net_trained.caffemodel')
for c,f in zip(values_celsius_test,values_farenheit_test):
	testnet.blobs['celsius'].data[0] = c
	testnet.forward()
	predicted_f = testnet.blobs['output'].data[0,0]
	print('Cesius: '+str(c)+'C, predicted: '+str(predicted_f)+' F, actual: '+str(f)+' F')
	error.append(abs(f-predicted_f))
print('Average error: '+str(np.array(error).mean())+' F')
\end{lstlisting}
Listing \ref{lst:celsius_farenheit_test} demonstrates how easily different inference precisions can be selected (line 6). The network will load the full-precision or already quantized weights (line 24), quantize them (if necessary) and run the inference according to the estimated quantization parameters.

The accuracy of the different inference precisions can be found in Section \ref{subsec:celsius_farenheit_acc}.

\section{MNIST}
\label{sec:mnist}
This example uses two fully connected layers to recognize hand-written digits from 1 to 9.

\begin{lstlisting}[caption={Creating the simple MNIST network}, label={lst:mnist_create}, language=Python]
import sys
sys.path.append('..')
from caffe_examples_setup import *

# Create a simple network with just one hidden layer and a flat 784 size input vector

# We create the network for float, half, int16 and int8
data_types = [caffe.data_type.CAFFE_HALF, caffe.data_type.CAFFE_FLOAT,
caffe.data_type.CAFFE_INT8_QUANTIZED, caffe.data_type.CAFFE_INT16_QUANTIZED]
data_types_names = ['half', 'float', 'int8', 'int16']

for data_type in zip(data_types, data_types_names):
	net = caffe.NetSpec()
	net.mnist_image = L.Input(input_param=dict(shape=dict(dim=[1,1,1,784])), ntop=1)
	net.label = L.Input(input_param=dict(shape=dict(dim=[1,1,1,1])), ntop=1)
	
	net.hidden_layer = L.InnerProduct(net.mnist_image,
									  bottom_data_type = data_type[0],
									  compute_data_type = data_type[0],
									  top_data_type = data_type[0],
									  inner_product_param = dict(
									  num_output = 30,
									  weight_filler = dict(type='xavier'),
									  bias_filler = dict(type='constant', value=0.0)))
	net.output_layer = L.InnerProduct(net.hidden_layer,
									  bottom_data_type = data_type[0],
									  compute_data_type = data_type[0],
									  top_data_type = data_type[0], 
									  inner_product_param = dict(
									  num_output = 10,
									  weight_filler = dict(type='xavier'),
									  bias_filler = dict(type='constant', value=0.0)))
	
	net.loss = L.SoftmaxWithLoss(net.output_layer, net.label,include=dict(phase=0))
	net.pred = L.Softmax(net.output_layer, include=dict(phase=1))

	protonet = net.to_proto()
	protonet.name = 'net'
	with open(protonet.name + '_' + data_type[1] + '.prototxt', 'w') as f:
		print(protonet, file=f)
\end{lstlisting}

\begin{lstlisting}[caption={Training the simple MNIST network}, label={lst:mnist_train}, language=Python]
import sys
sys.path.append('..')
from caffe_examples_setup import *

# Choose the precision (half, float, int16 or int8).
precision = 'float'

# Load the data
f = open('../data/mnist.pkl', 'rb')
training_data, validation_data, test_data = cPickle.load(f, encoding='latin1')
f.close()

# Create a solver with a few typical parameters
# The solver will perform SGD on our data
solver_config = caffe.SolverParameter()
solver_config.train_net = 'net_' + precision + '.prototxt'
solver_config.base_lr = 0.01
solver_config.momentum = 0.99
solver_config.weight_decay = 0.0001
solver_config.lr_policy = 'inv'
solver_config.gamma = 0.0001
solver_config.power = 0.75
solver_config.max_iter = 16000
solver_config.snapshot = 4000
solver_config.snapshot_prefix = 'net'
solver_config.type = 'Adam'
solver_config.display = 100

# Do the training
losses = []

plt.ion()
plot_obj, = plt.plot(losses)
ax = plt.gca()
plt.show()
plt.pause(0.001)

solver = caffe.get_solver(solver_config)
for i in range(0, solver_config.max_iter):
	# Pick a random sample for training
	k = random.randint(0,len(training_data[0])-1)
	# Load the sample into the network
	solver.net.blobs['mnist_image'].data[:] = np.reshape(training_data[0][k],(784)).astype(float)/255.0
	solver.net.blobs['label'].data[0] = training_data[1][k]
	# Train one step
	loss = solver.step(1)
	# Display the learning progress every 20 steps
	if (i % 100 == 0):
		losses.append(loss)
		plot_obj.set_data(range(0,len(losses)), losses)
		ax.relim()
		ax.autoscale_view(True, True, True)
		plt.draw()
		plt.pause(0.001)


# Run a few test steps to observe the value ranges (for quantization)
error = 0
testnet = caffe.Net(str('net_' + precision + '.prototxt'), caffe.TEST, weights='net_iter_'+str(solver_config.max_iter)+'.caffemodel')
# Enable quantizer observation
testnet.quant_mode = caffe.quantizer_mode.CAFFE_QUANT_OBSERVE
for k in range(0,len(validation_data[0])):
	testnet.blobs['mnist_image'].data[:] = np.reshape(validation_data[0][k],(784)).astype(float)/255.0
	testnet.forward()
	if (k % 100 == 0):
		print(k)
# Store the network parameters, including obtain quantizer information
testnet.save('net_trained.caffemodel')
print("Done.")
\end{lstlisting}

\begin{lstlisting}[caption={Testing the simple MNIST network}, label={lst:mnist_test}, language=Python]
import sys
sys.path.append('..')
from caffe_examples_setup import *

# Choose the precision (half, float, int16 or int8)
precision = 'float'

# Load the data
f = open('../data/mnist.pkl', 'rb')
training_data, validation_data, test_data = cPickle.load(f, encoding='latin1')
f.close()

# Test how accurate the network has learned it's task
error = 0
testnet = caffe.Net(str('net_' + precision + '.prototxt'), caffe.TEST, weights='net_trained.caffemodel')
for k in range(0,len(validation_data[0])):
	testnet.blobs['mnist_image'].data[:] = np.reshape(validation_data[0][k],(784)).astype(float)/255.0
	testnet.forward()
	predicted_number = np.argmax(testnet.blobs['pred'].data[:])
	print('Predicted: '+str(predicted_number)+', actual: '+str(validation_data[1][k]))
	if not (predicted_number == validation_data[1][k]):
		error += 1
print('Errors: '+str(error)+' of '+str(len(validation_data[0]))+' ('+str(100.0-100.0*((float)(error)/(float)(len(validation_data[0]))))+'% accuracy)')
\end{lstlisting}

The accuracy of the different inference precisions can be found in Section \ref{subsec:mnist_acc}.
\chapter{ImageNet}
\label{ch:imagenet}
To assess the memory, storage and compute requirements of our methods, we chose the well-established AlexNet/ImageNet \cite{AlexNet2012} on the 1000-way image classification task ILSVRC2012. The input size of the network is an RGB image with $227\times227$ pixels.

\section{ImageNet Configuration}
\label{sec:imagenet_config}

As a baseline, we used the standard AlexNet/ImageNet \cite{AlexNet2012} bundled with Caffe \cite{Caffe}.
It has a single compute path, defined by following operations:
\begin{itemize}
	\item Convolution (kernel size 11, stride 4, 96 feature maps) + ReLU
	\item Pooling (kernel size 3, stride 2) + LRN
	\item Convolution (kernel size 5, pad 2, group 2, 256 feature maps) + ReLU
	\item Pooling (kernel size 3, stride 2) + LRN
	\item Convolution (kernel size 3, pad 1, 384 feature maps) + ReLU
	\item Convolution (kernel size 3, pad 1, group 2, 384 feature maps) + ReLU
	\item Convolution (kernel size 3, pad 1, group 2, 256 feature maps) + ReLU
	\item Pooling (kernel size 3, stride 2)
	\item Fully connected (4096 feature maps) + ReLU + Dropout
	\item Fully connected (4096 feature maps) + ReLU + Dropout
	\item Fully connected (1000 feature maps) + Softmax
\end{itemize}

The network graph can be found in Appendix \ref{sec:app_imagenet}.

\newpage
\section{ImageNet-MOE Configuration}
\label{sec:imagenet_moe_config}

We build the MOE network configuration according to the method description in Section \ref{sec:mixture_of_experts}.

\subsection{MOE Main Network}
\begin{itemize}
	\item Convolution (kernel size 11, stride 4, 48 feature maps) + ReLU
	\item Pooling (kernel size 3, stride 2) + LRN
	\item Mixture-of-Experts (16 experts, 4 experts per sample, 2048 feature maps) + ReLU + Dropout
	\item Fully connected (1000 feature maps) + Softmax
\end{itemize}

The main network includes the first and last few layers of the original ImageNet, with slightly altered feature map counts.

\subsection{MOE Gating Network}
\label{subsec:imagenet_moe_gating}
Our gating network is substantially smaller and cheaper to compute than the expert network, because it is usually enough to get approximate hints out of the gating network. Smaller networks are also easier to train in this case, since the regularization loss and noise (see Section \ref{sec:mixture_of_experts}) complicate the training.

\begin{itemize}
	\item Convolution (kernel size 5, pad 2, group 2, 64 feature maps) + ReLU
	\item Pooling (kernel size 3, stride 2) + LRN
	\item Fully connected (128 feature maps) + ReLU
	\item Fully connected (16 feature maps)
\end{itemize}

In more complex, hierarchical tasks, it would be possible to additionally train the gating network with a task such as predicting the class an object belongs to, while the whole network would classify the exact object type.

\subsection{MOE Expert Network}
\label{subsec:imagenet_moe_expert}
The expert network in our model is repeated 16 times, with each layer, except the output layer, having four times fewer output feature maps compared to the original AlexNet. This makes each expert network approximately 16 times cheaper to compute, since the number of computations depends on the product of input- and output-feature maps of each layer. Per forward pass, the MOE layer chooses 4 of 16 experts, reducing computation requirements four times. The mixture-of-experts layer and the gating network add some overhead again, however.

\begin{itemize}
	\item Convolution (kernel size 5, pad 2, group 2, 64 feature maps) + ReLU
	\item Pooling (kernel size 3, stride 2) + LRN
	\item Convolution (kernel size 3, pad 1, 96 feature maps) + ReLU
	\item Convolution (kernel size 3, pad 1, group 2, 96 feature maps) + ReLU
	\item Convolution (kernel size 3, pad 1, group 2, 64 feature maps) + ReLU
	\item Pooling (kernel size 3, stride 2)
	\item Fully connected (1024 feature maps) + ReLU + Dropout
	\item Fully connected (2048 feature maps)
\end{itemize}

The expert network includes the middle hidden layers of the original ImageNet, which are not included in the main MOE network. Combining the two parts results in a forward-backward path with equivalent feature map sizes to the original ImageNet.

The network graph can be found in Appendix \ref{sec:app_imagenet_moe}.
\chapter{Benchmarks}
\label{ch:benchmarks}
In order to assess performance and accuracy of both the mixture-of-experts networks and mixed-precision computations, we used the classic ImageNet/AlexNet \cite{AlexNet2012} as a case study (see Chapter \ref{ch:imagenet}).

\section{Devices}
\label{sec:devices}
We benchmarked following devices:
\begin{table}[H]
	\centering
	\begin{tabular}{p{0.23\textwidth}|p{0.21\textwidth}p{0.21\textwidth}p{0.21\textwidth}p{0.21\textwidth}}
		\hline
		Device Name & AMD Vega FE\newline \cite{VegaWhitepaper} & AMD RX 480\newline \cite{PolarisWhitepaper} & nVidia GTX 1080\newline \cite{GTX1080Whitepaper} \\\hline\hline
		Compute Units & 64 (4096) & 36 (2304) & 20 (2560)\\
		Memory [\SI{}{GiB}] & 16 & 8 & 8\\
		TFLOPS FP16 & 26.2 & 5.8$^{(2)}$ & 0.6$^{(3)}$\\
		TFLOPS FP32 & 13.1 & 5.8 & 8.9\\
		TOPS INT8 & 52.4$^{(1)}$ & 5.8 & 35.6$^{(4)}$\\
		TOPS INT16 & 26.2$^{(1)}$ & 5.8 & 17.8$^{(4)}$\\
		\hline
	\end{tabular}\\\bigskip
	\begin{tabular}{p{0.23\textwidth}|p{0.21\textwidth}p{0.21\textwidth}p{0.21\textwidth}p{0.21\textwidth}}
		\hline
		Device Name & nVidia GT 1030 \cite{GTX1080Whitepaper} & Intel Iris Pro 540 \cite{IntelIrisProGraphics, intelark} & ARM Mali T764 \cite{asustinkerboard,ARMMaliGPUOpenCL} \\\hline\hline
		Compute Units & 3 (384) & 2$\times$3 (48) & 4 (64)\\
		Memory [\SI{}{GiB}] & 2 & 5 & 0.25\\
		TFLOPS FP16 & 0.018$^{(3)}$ & 1.504$^{(5)}$ & 0.16\\
		TFLOPS FP32 & 1.127 & 0.752 & 0.08\\
		TOPS INT8 & 4.508 & 0.752 & 0.32\\
		TOPS INT16 & 2.254 & 0.752 & 0.16\\
		\hline
	\end{tabular}
	\caption{Devices used for benchmarks and their features.\newline $^{(1)}$ Only fast integer for some operators, which are not useful to Caffe inference.\newline$^{(2)}$ Uses FP32 path for computation and FP16 for storage only.\newline $^{(3)}$ Has a FP16 code path for computation, but is limited at $\sfrac{1}{64}$ FP32 speed.\newline$^{(4)}$ Using nVidias DP4 and DP2 instructions \cite{cudadp4dp2}.\newline$^{(5)}$ Not working due to OpenCL driver issues on Intel Beignet.}
	\label{tab:devices}
\end{table}

\section{Memory and Storage Consumption}
\label{sec:memory_consumption}
\subsection{Storage}
\label{subsec:storage}

\begin{table}[H]
	\centering
	\begin{tabular}{lrr}
		\hline
		Network & ImageNet & ImageNet-MOE\\\hline
		FP32 & \SI{243.9}{MiB} & \SI{309.9}{MiB}\\
		FP16 & \SI{121.9}{MiB} & \SI{155.0}{MiB}\\
		INT16 & \SI{121.9}{MiB} & \SI{155.0}{MiB}\\
		INT8 & \SI{61.0}{MiB} & \SI{77.5}{MiB}\\
		\hline
	\end{tabular}
	\caption{Network parameter storage size.}
	\label{tab:storage} 
\end{table}

\begin{figure}[H]
	\centering
	\includegraphics[scale=0.8]{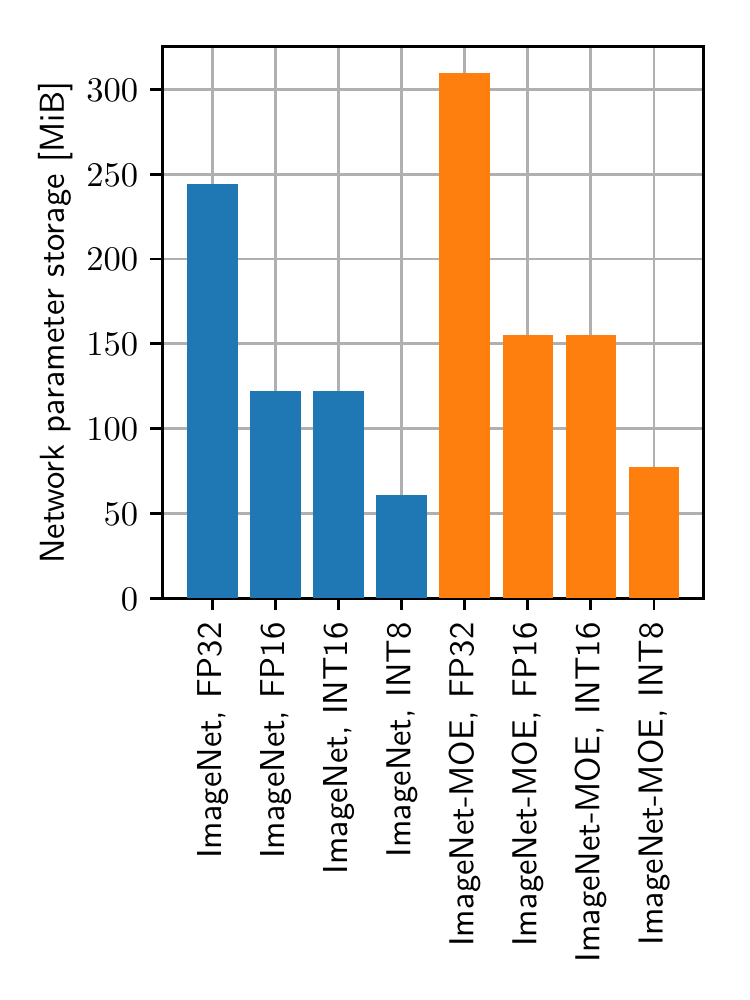}
	\caption{Network parameter storage size.}
	\label{fig:storage} 
\end{figure}

Caffe now has the possibility to also store the trained weights in reduced precision. As expected, storage requirements drop linearly with the number of bytes per weight. With FP32, 4 bytes per weight are consumed, while with the lowest precision, INT8, it is only one byte. The mixture-of-experts variant of ImageNet uses slightly more memory (27\%), which can mostly be attributed to the gating network and the increased number of weights when transiting from the main network into the expert networks and back again.

The quantization parameters, which are always stored as FP32, do not add significantly to the storage requirements.

\subsection{Memory}

For memory consumption, we measured only the GPU memory, while ignoring the overhead memory on the CPU. All memory related tests have been executed for a batch size of 512 images.

\begin{table}[H]
	\centering
	\begin{tabular}{l|rrrr}
		\hline
		Network & ImageNet & ImageNet & ImageNet-MOE & ImageNet-MOE\\\hline\hline
		Reduced memory & no & yes & no & yes\\
		CUDA FP32 & \SI{3351}{MiB} & \SI{2281}{MiB} & \SI{6753}{MiB} & \SI{3991}{MiB}\\
		CUDA FP16 & \SI{2389}{MiB} & \SI{1407}{MiB} & \SI{1409}{MiB} & \SI{1209}{MiB}\\
		CUDA INT16 &\SI{2391}{MiB} & \SI{1409}{MiB} & \SI{4535}{MiB} & \SI{2631}{MiB}\\
		CUDA INT8 & \SI{1815}{MiB} & \SI{1209}{MiB} & \SI{3361}{MiB} & \SI{2061}{MiB}\\
		OpenCL FP32 & \SI{3275}{MiB} & \SI{2047}{MiB} & \SI{6211}{MiB} & \SI{3449}{MiB}\\
		OpenCL FP16 & \SI{2341}{MiB} & \SI{1379}{MiB} & \SI{4329}{MiB} & \SI{2432}{MiB}\\
		OpenCL INT16 & \SI{2343}{MiB} & \SI{1361}{MiB} & \SI{4361}{MiB} & \SI{2457}{MiB}\\
		OpenCL INT8 & \SI{1771}{MiB} & \SI{1165}{MiB} & \SI{3187}{MiB} & \SI{1887}{MiB}\\
		\hline
	\end{tabular}
	\caption{Network memory consumption.}
	\label{tab:memory} 
\end{table}

\label{subsec:memory}
\begin{figure}[H]
	\centering
	\includegraphics[scale=0.8]{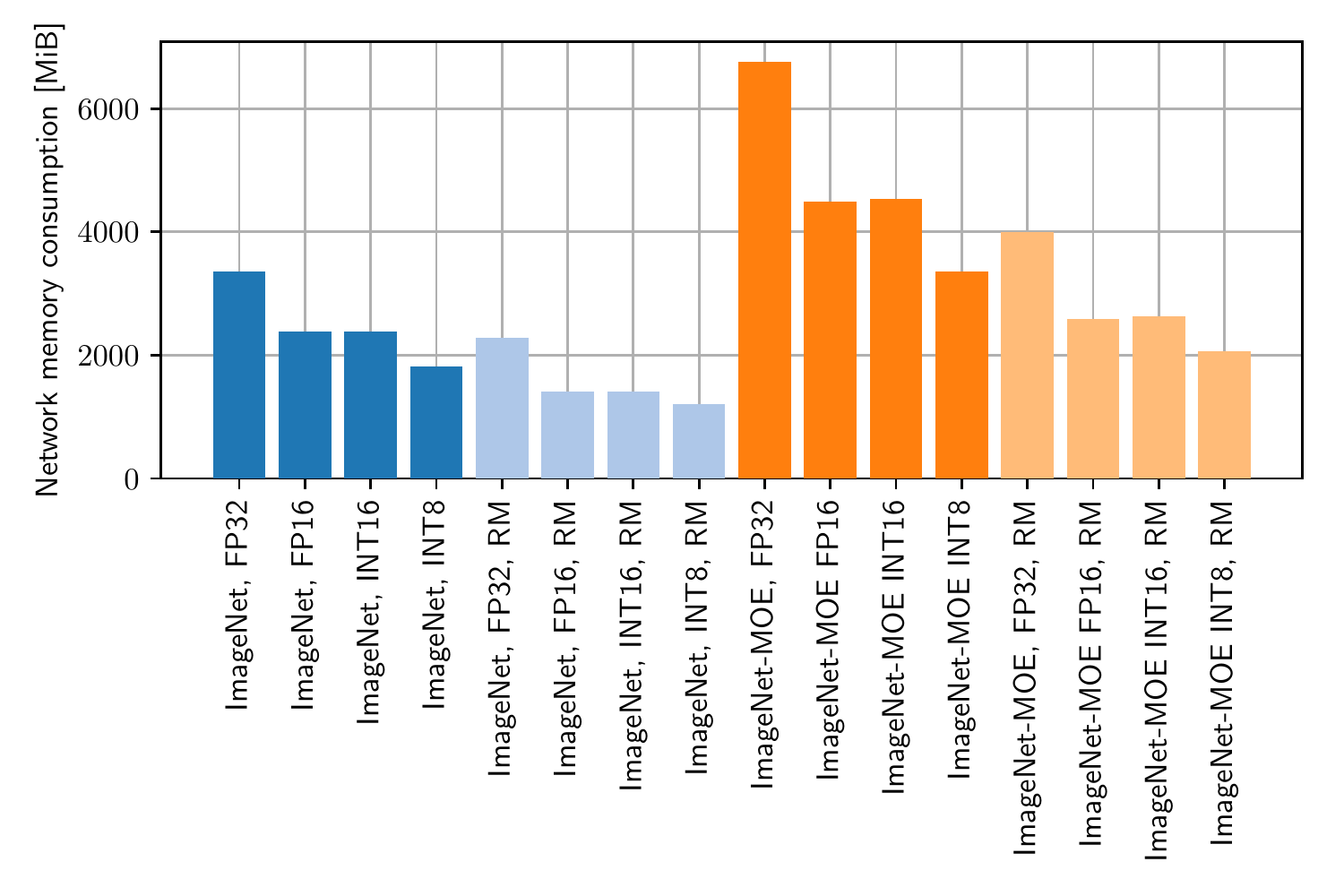}
	\caption{Network memory consumption using the CUDA backend.}
	\label{fig:memory_cuda} 
\end{figure}

Using the lowest precision, INT8, reduces the memory consumption by \SI{54}{\%} compared to FP32. Since we use mixed-precision, some layers such as the LRN layer are still executed at full precision, even when using INT8. This means we get less than linear improvement.

\begin{figure}[H]
	\centering
	\includegraphics[scale=0.8]{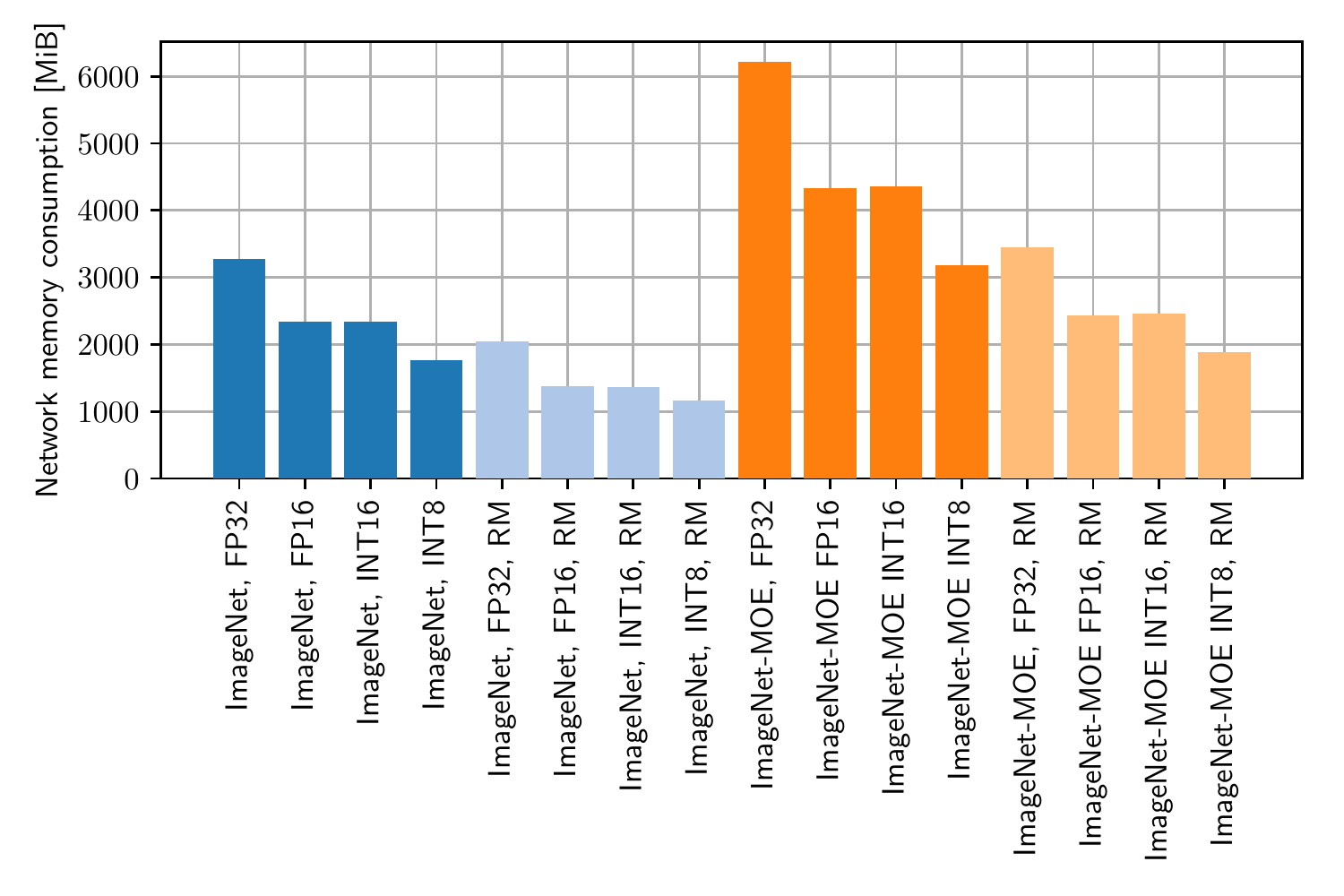}
	\caption{Network memory consumption using the OpenCL backend.}
	\label{fig:memory_opencl} 
\end{figure}

With an additional technique, which reuses the blobs in the network graph as often as possible (denoted RM in Figures \ref{fig:memory_cuda} and \ref{fig:memory_opencl}), we can reduce the memory consumption by an additional \SI{68}{\%}. The downside of reusing memory is that inspection of intermediate results in the network is impossible, but this is not a problem in typical inference applications. When debugging a network, the memory sharing can easily be disabled. During training, no blobs in the network are allowed to be overwritten, since they are needed for gradient computations. Therefore, the reduced memory option is always disabled during training.

In total, we claim up to $3.29\times$ less memory consumption, using a combination of low-precision and reduced-memory inference.

OpenCL typically uses less memory than CUDA, which we tracked down to pre-allocated memory inside the external libraries that Caffe makes use of. Such temporary memory can be used to store intermediate results of reduction, convolution and other operators that require global working memory.

\section{Inference Throughput}
\label{sec:throughput}

\subsection{ImageNet}
\label{subsec:imagenet_throughput}

\begin{table}[H]
	\centering
	\begin{tabular}{llr|rrrr}
		\hline
		GPU & Backend & Batch & FP32 & FP16 & INT16 & INT8\\\hline\hline
		nVidia GTX 1080 & CUDA & 512 & \SI{510}{ms} & \SI{22333}{ms} & \SI{2350}{ms} & \SI{565}{ms}\\
		nVidia GTX 1080 & OpenCL & 512 & \SI{728}{ms} & n/a & \SI{1425}{ms} & \SI{715}{ms}\\
		nVidia GT 1030 & CUDA & 64 & \SI{353}{ms} & \SI{18014}{ms} & \SI{1595}{ms} & \SI{437}{ms}\\
		nVidia GT 1030 & OpenCL & 64 & \SI{542}{ms} & n/a & \SI{1258}{ms} & \SI{594}{ms}\\
		AMD Vega FE & OpenCL & 512 & \SI{863}{ms} & \SI{380}{ms} & \SI{828}{ms} & \SI{931}{ms}\\
		AMD RX 480 & OpenCL & 512 & \SI{1675}{ms} & \SI{1014}{ms} & \SI{1863}{ms} & \SI{2038}{ms}\\
		Intel Iris Pro 540 & OpenCL & 64 & \SI{1209}{ms} & n/a & \SI{24737}{ms} & \SI{9926}{ms}\\
		ARM Mali T764 & OpenCL & 4 & \SI{4321}{ms} & \SI{2461}{ms} & \SI{4045}{ms} & \SI{3606}{ms}\\\hline
	\end{tabular}
	\caption{ImageNet inference time.}
	\label{tab:imagenet_timing} 
\end{table}

\begin{figure}[H]
	\centering
	\includegraphics[width=\textwidth]{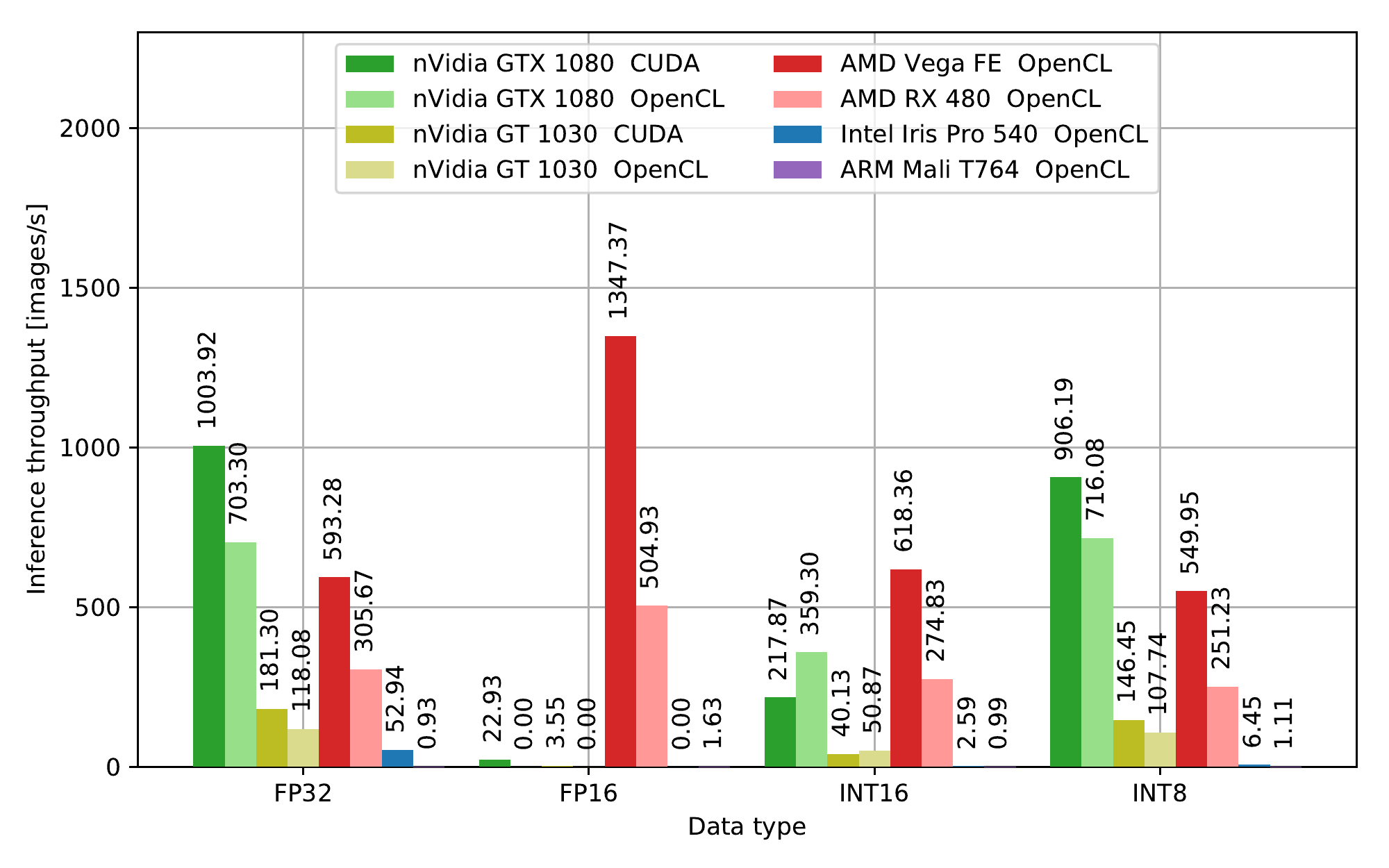}
	\caption{ImageNet throughput.}
	\label{fig:imagenet_throughput} 
\end{figure}

Using FP16 increases the throughput on both AMD GPUs and on the ARM Mali T764 GPU, however, not by the same amount. Because the AMD Vega FE and Mali GPU can execute twice the amount of FP16 operations compared to FP32 (see Table \ref{tab:devices}), their throughput increases by \SI{128.8}{\%} and \SI{75.3}{\%} respectively. The AMD RX 480 card, which can do FP16 computations, but uses FP32 internally, gains \SI{65}{\%} through memory bandwidth savings on the global, local and register memory. For nVidia and Intel GPUs, the FP16 data type is not useful.\newline

The integer quantized types are, at large, not useful to increase the inference speed on any GPU we tested. Since the speed of integer types is consistent on the AMD GPUs, it may still be a viable option if memory is the limiting factor (see Section \ref{sec:memory_consumption})

nVidia GPUs use DP4A and DP2A \cite{cudadp4dp2} instructions to accelerate INT8 computations, but the compute and memory overhead of the quantized types negate any performance gains in our implementation. We confirmed that the DP4A and DP2A instructions are actually compiled into the compute kernels using assembly code analysis. Additional kernel tuning and probably hand-tuned algorithms would be required to reach higher throughputs.

Since INT8 quantized computation also requires INT16 and INT32 operations (see Section \ref{subsec:quantized_operations}), GPUs like the ARM Mali T764 and AMD Vega FE, which could execute some pure INT8 operations (addition, subtraction, quad-absolute-sum-of-differences) at faster speeds, do not profit from the lower precision inference paths. Intrinsics that perform fused-multiply-add from INT8 to INT32 on vector types are likely required to increase performance, but are not implemented in the hardware.

\begin{figure}[H]
	\centering
	\includegraphics[width=\textwidth]{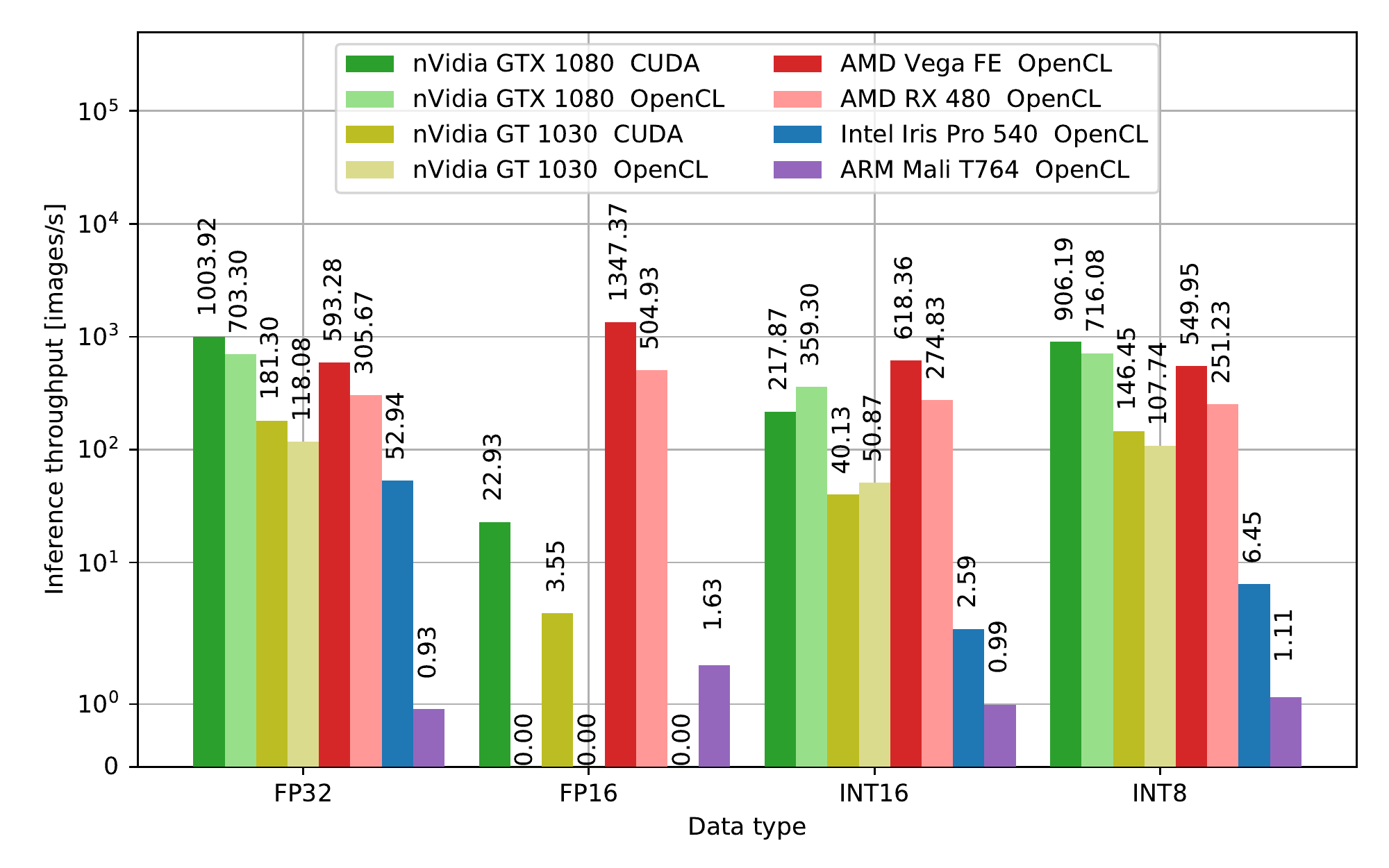}
	\caption{ImageNet throughput (log scale).}
	\label{fig:imagenet_throughput_log} 
\end{figure}

\subsection{ImageNet-MOE}
\label{subsec:imagenet_moe_throughput}

\begin{table}[H]
	\centering
	\begin{tabular}{llr|rrrr}
		\hline
		GPU & backend & Batch & FP32 & FP16 & INT16 & INT8\\\hline\hline
		nVidia GTX 1080 & CUDA & 512 & \SI{587}{ms} & \SI{79770}{ms} & \SI{7743}{ms} & \SI{1626}{ms}\\
		nVidia GTX 1080 & OpenCL & 512 & \SI{1541}{ms} & n/a & \SI{4776}{ms} & \SI{2342}{ms}\\
		nVidia GT 1030 & CUDA & 64 & \SI{385}{ms} & \SI{63934}{ms} & \SI{5136}{ms} & \SI{1267}{ms}\\
		nVidia GT 1030 & OpenCL & 64 & \SI{1124}{ms} & n/a & \SI{3884}{ms} & \SI{1824}{ms}\\
		AMD Vega FE & OpenCL & 512 & \SI{1353}{ms} & \SI{1017}{ms} & \SI{2433}{ms} & \SI{2797}{ms}\\
		AMD RX 480 & OpenCL & 512 & \SI{2453}{ms} & \SI{2373}{ms} & \SI{5002}{ms} & \SI{5881}{ms}\\
		Intel Iris Pro 540 & OpenCL & 64 & \SI{3600}{ms} & n/a & \SI{41403}{ms} & \SI{14789}{ms}\\
		ARM Mali T764 & OpenCL & 4 & \SI{2422}{ms} & \SI{1435}{ms} & \SI{5837}{ms} & \SI{6096}{ms}\\\hline
	\end{tabular}
	\caption{ImageNet-MOE inference time.}
	\label{tab:imagenet_moe_timing} 
\end{table}

\begin{figure}[H]
	\centering
	\includegraphics[width=\textwidth]{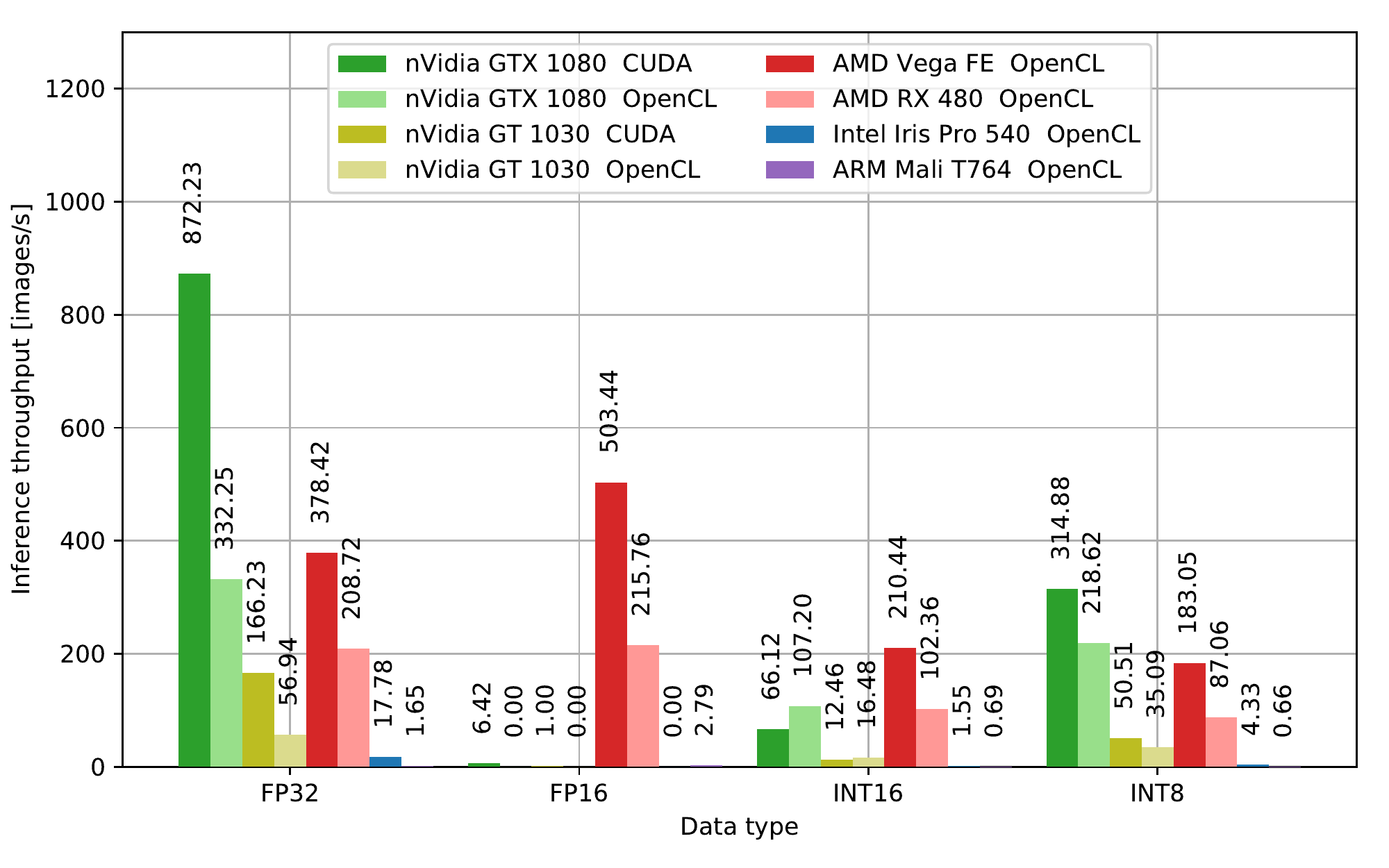}
	\caption{ImageNet-MOE throughput.}
	\label{fig:imagenet_moe_throughput} 
\end{figure}

Performance on our ImageNet-MOE network is consistently slower on all GPUs except the ARM Mali T764. This was an expected outcome, since the larger GPUs profit from executing a network for a large batch at once. For the GPUs using a batch-size of 512 and for the nVidia GT 1030 (see Table \ref{tab:imagenet_moe_timing}), we configured the MOE layer so that all experts are always computed for the whole batch. Executing the experts only for the necessary samples resulted in even lower throughputs, because the individual computations are too small to utilize the whole GPU.

We tried to mitigate the performance reduction problem by at least executing all experts in parallel, using up to 8 streams/queues. But because each operator in the expert networks also have fewer operations, due to reduced input- and output-feature maps (see Section \ref{sec:imagenet_moe_config}), getting the same utilization as a normal ImageNet is not possible. We therefore conclude that mixture-of-experts only become useful on large GPUs when the individual experts are expensive to compute and can be executed for a large batch of samples using the same experts at once.

An alternative use-case is low-latency inference, where for example a live-stream from a camera has to be passed through a neural network, making large batches inherently impossible. In this case, using MOE networks may also make sense on large GPUs.

\begin{figure}[H]
	\centering
	\includegraphics[width=\textwidth]{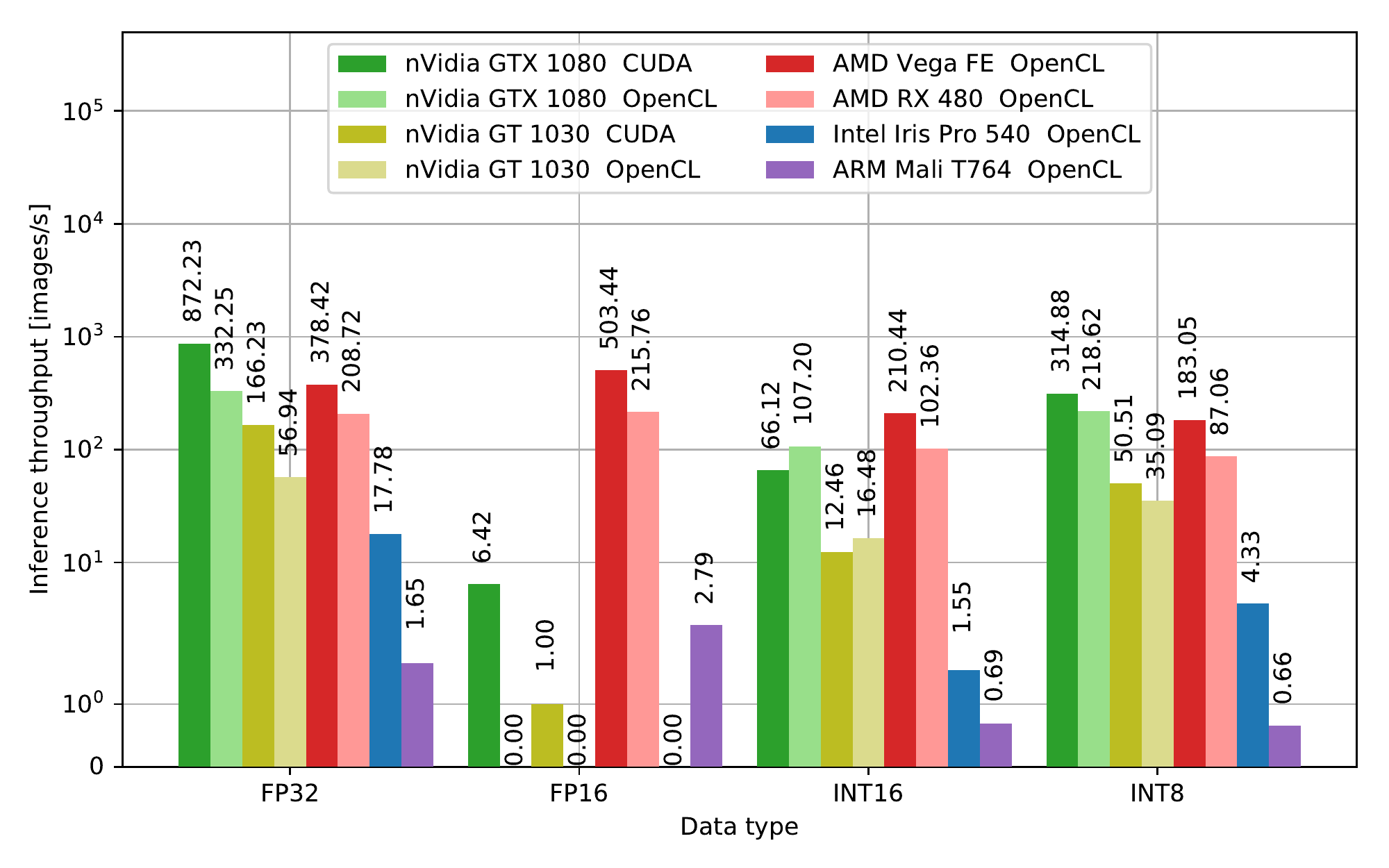}
	\caption{ImageNet-MOE throughput (log scale).}
	\label{fig:imagenet_moe_throughput_log} 
\end{figure}

On the Intel Iris Pro 540 and ARM Mali T764, only computing the necessary experts selected by the gating network per-sample (see Section \ref{sec:mixture_of_experts}) results in higher inference speed (see Figures \ref{fig:imagenet_moe_throughput} and \ref{fig:imagenet_moe_throughput_log}).

For the ARM Mali T764, the inference speed increases \SI{77}{\%} compared to the normal ImageNet at FP32, and \SI{71}{\%} at FP16.

Using a combination of FP16 inference and MOE technique, we achieve the claimed $3.01\times$ throughput, compared to normal ImageNet FP32 inference.

\newpage
\section{Inference Accuracy}
\label{sec:inference_accuracy}

\subsection{Celsius-Farenheit}

We included the accuracy of the celsius-farenheit example (see Section \ref{sec:celsius_farenheit_example}) because on this example, the accuracy results can very easily be explained.
\label{subsec:celsius_farenheit_acc}
\begin{table}[H]
	\centering
	\begin{tabular}{l|cccc}
		\hline
		& FP32 & FP16 & INT16 & INT8\\
		\hline
		\hline
		Error [$\Delta^{\circ}$F] & 2.51804 & 2.26494 & 2.42341 & 6.93617\\
		\hline
	\end{tabular}
	\caption{Celsius to farenheit conversion error for different precisions.}
	\label{tab:celsius_farenheit_acc}
\end{table}
We see in Table \ref{tab:celsius_farenheit_acc} that FP32, FP16 and INT16 have negligible difference in accuracy. This is explained by each of these data types being able to sufficiently represent the range of values for both the input in $^{\circ}C$ in the range of $[-273.0, 1000.0)$ as well as the output in $^{\circ}F$ in the range of $[-394.6,1832)$. The same is trivially true for the learned weight ($a=1.8$) and bias ($b=32.0$) parameters. For INT16, we have 65'535 binned values (see Section \ref{sec:quantization}), which would allow the error to be as low as $0.017\,^{\circ}F$ with sufficient training.

INT8, on the other hand, only has 255 binned values. This restricts the error to be greater or equal to $4.3658\,^{\circ}F$, which is why INT8 performs worse than the other data types. This should be kept in mind when designing neural networks. Especially the input and output stages might benefit from higher precisions in order to accurately represent the value domains.

\subsection{MNIST}
\label{subsec:mnist_acc}
\begin{table}[H]
	\centering
	\begin{tabular}{l|cccc}
		\hline
		& FP32 & FP16 & INT16 & INT8\\
		\hline
		\hline
		Top-1 Accuracy [\si{\%}] & 85.26 & 85.30 & 85.27 & 85.36\\
		\hline
	\end{tabular}
	\caption{MNIST accuracy.}
	\label{tab:mnist_acc}
\end{table}

We recognize that MNIST \cite{lecun-mnisthandwrittendigit-2010} can be predicted to a much higher accuracy with models such as LeNet \cite{LeNet1998}. However, for simplicity, as an usage example of mixed-precisions, we demonstrate a simple network with only two fully connected layers (see Section \ref{sec:mnist}). The more complex LeNet example can be found online \cite{CaffeExamples}.
All data types achieve the same accuracy. Since the input domain is pixel values in $[0, 255]$ and the output is discrete in $[0, 9]$, this problem does not suffer from domain binning restrictions by quantization, unlike converting celsius to farenheit values (see Section \ref{subsec:celsius_farenheit_acc}). Another aspect is that the model has more weights, making the precision of each single weight less important.

\subsection{ImageNet}
\label{subsec:imagenet_acc}

\begin{table}[H]
	\centering
	\begin{tabular}{l|cccc}
		\hline
		& FP32 & FP16 & INT16 & INT8\\
		\hline
		\hline
		Top-1 accuracy (train) [\si{\%}] & 79.113 & 79.059 & 54.566 & 59.840\\
		Top-5 accuracy (train) [\si{\%}] & 93.887 & 93.852 & 77.262 & 83.277\\
		Top-1 accuracy (test) [\si{\%}] & 56.875 & 56.813 & 42.690 & 44.750\\
		Top-5 accuracy (test) [\si{\%}] & 79.980 & 79.973 & 66.320 & 70.043\\
		\hline
	\end{tabular}
	\caption{Original ImageNet accuracy}
	\label{tab:imagenet_acc}
\end{table}

The standard pre-trained ImageNet reaches a test-set accuracy of up to \SI{56.875}{\%}. Using the lower precision inference FP16 is a good choice, since it does not drop the accuracy significantly, does not require fine tuning and reaches over \SI{100}{\%} increased throughput over FP32 on some devices (see Section \ref{subsec:imagenet_throughput}).

Using INT8 and INT16 reduces accuracy up to \SI{12}{\%} without fine-tuning the network with pseudo-quantization (see Section \ref{sec:quantization}).

\subsection{ImageNet-MOE}
\label{subsec:imagenet_moe_acc}
\begin{table}[H]
	\centering
	\begin{tabular}{l|cccc}
		\hline
		& FP32 & FP16 & INT16 & INT8\\
		\hline
		\hline
		Top-1 accuracy (train) [\si{\%}] & 66.804 & 66.734 & 12.805 & 53.410\\
		Top-5 accuracy (train) [\si{\%}] & 86.933 & 86.906 & 25.531 & 78.363\\
		Top-1 accuracy (test) [\si{\%}] & 36.125 & 36.078 & 8.980 & 30.305\\
		Top-5 accuracy (test) [\si{\%}] & 59.473 & 59.468 & 19.359 & 53.355\\
		\hline
	\end{tabular}
	\caption{ImageNet-MOE accuracy}
	\label{tab:imagenet_MOE_acc}
\end{table}

The ImageNet-MOE model only reached \SI{36.125}{\%} on the ILSVRC validation set, however the training set scores hint that the network is fully capable of learning the task. It is likely that our choice of gating selector (see Section \ref{sec:mixture_of_experts}) is not optimal yet. It is difficult to tune the amount of noise and the learning rate of the regularizer so that the experts learn optimally. We still observed some experts being disabled completely, while one expert was used for every sample. This reduces the learning capability of the network drastically. A more elaborated gating selector \cite{Shazeer2017} might also improve accuracy.

Interestingly, for both ImageNet and ImageNet-MOE, the INT16 score is lower than on INT8, hinting at possible integer value casting problems either in the implementation or the compilers. Since there are not many reasons to use the INT16 data type on the devices and networks we tested, it is probably not a good choice for any real use-case, except for toy examples like the celsius-farenheit conversion example (see Section \ref{subsec:celsius_farenheit_acc} and \ref{sec:celsius_farenheit_example}) or on embedded devices with no hardware floating point capabilities.
\chapter{Conclusion}
\label{ch:conclusion}

\section{Implications}
\label{sec:implications}
We successfully implemented a more versatile backend for Caffe, allowing to use a large variety of compute libraries, hardware backends, devices and data types. The added flexibility will allow more use cases for Caffe, and position it as a go-to choice in applications that have to be distributed over a large variety of different hardware.

Our new models, compute paths and improvements to the memory system of Caffe also enable up to $3.29\times$ less memory usage, while increasing inference speed up to $3.01\times$ on certain devices. This can make the difference of being able to use neural networks on low-power devices or not.

\section{Difficulties Encountered}
\label{sec:difficulties}
While implementing quantizers (see Section \ref{sec:quantization}) and device abstraction (see Section \ref{sec:quantization}) was challenging from a software engineering perspective, the most annoying and difficult issues were bugs in the drivers of GPUs from all vendors. Until the latest updates (May 2018) were available from both AMD (ROCm 1.8) and nVidia (CUDA 9.2), certain operators (FP16 and INT8) did not work at their full speed, and certain kernels did not compile at all. This was most likely due to immature compilers that did not handle the OpenCL and CUDA kernels optimally for the latest GPU architectures. For the Intel GPU, no update was released yet to remedy their bugs on the FP16 implementation (see Table \ref{tab:devices}).

Finally, finding the right hyper-parameters to configure the MOE layer (see Section \ref{sec:mixture_of_experts}) and train the modified ImageNet/AlexNet \cite{AlexNet2012} was frustrating, since it only becomes apparent if a model will converge or not at a rather late stage in training, typically after 5 to 8 hours (100'000 iterations). Only when training a seemingly non-working configuration from start to finish once, did it suddenly begin to work. It seems like neural networks, when becoming increasingly complex, such as adding divergent code paths, remain black boxes \cite{Benitez1997} and definitely have a personality on their own.

Unfortunately, the very extensive scope of this project, unusual circumstances and computer driver problems made it difficult to arrive at the desired results in a reasonable time-frame.

\section{Reproducibility of Results}
\label{sec:reproducibility}
The results obtained in this project can be reproduced by the use of the following software pipeline, using CUDA or OpenCL hardware equivalent to the hardware used in this project.

Repositories belonging to the OpenCL Caffe Project \textit{}:
\begin{itemize}
	\item Caffe OpenCL development branch \cite{Caffe}
	\subitem URL: \url{https://github.com/naibaf7/caffe}
	\item Caffe OpenCL release branch \cite{Caffe}
	\subitem URL: \url{https://github.com/BVLC/caffe/tree/opencl}
	\item Caffe Examples \cite{CaffeExamples}
	\subitem URL: \url{https://github.com/naibaf7/opencl_caffe_examples}
\end{itemize}

\section{Outlook}
\label{sec:outlook}
While the new Caffe code provides a solid basis for future developments, allowing to now easily program for both OpenCL and CUDA, without divergent code paths, there are still many points that need improvement:

\begin{itemize}
	\item Performance on ARM devices and mobile GPUs is still not optimal. Additional hardware-specific libraries such as the ARM compute library \cite{armcompute} should offer improvement over the current approach of using LibDNN as a fallback library for all devices.
	\item OpenCL performance for desktop GPUs can be improved further. Adding AMD's new HIP \cite{amdhip} compute backend and accompanying compute libraries may help with performance. 
	\item The Python interface could now easily include a way to write GPU layers. This would allow the users to prototype GPU layers without the hassle of changing the Caffe core library and recompiling.
	\item MobileNets \cite{MobileNets2017} have been deemed suitable for low-power and embedded application. These could also be ported to Caffe and be optimized for lower precision, embedded GPUs, and could potentially be improved even further by applying mixed-precision and the mixture-of-experts pattern (see Section \ref{sec:mixture_of_experts}).
\end{itemize}

\newpage
\section{Final Words}
\label{sec:final_words}
This project, for now, concludes our development efforts on OpenCL Caffe. Through working on the OpenCL Caffe project since 2014 \cite{Tschopp2015, Tschopp2016}, I gained a full-stack development experience in the realm of deep learning. These efforts, as a cumulative product, enable more people to use deep learning effectively.
I was able to implement most of the originally planned features and to gain deep learning community interest in the project.

Continued hardware sponsoring and collaboration by Intel \cite{Intel} and AMD \cite{AMD} shows that such efforts to diversify the development of deep learning libraries continue to be highly important for industry, end-users and scientific progress.
%% \addtocontents{toc}{\protect\newpage}
%% \addtocontents{toc}{\protect\newpage}

\appendix

\chapter{Models}
\label{ch:app_models}
\section{Caffe-ImageNet}
\label{sec:app_imagenet}
\begin{figure}[H]
	\centering
	\begin{subfigure}{.5\textwidth}
		\centering
		\includegraphics[scale=1.7]{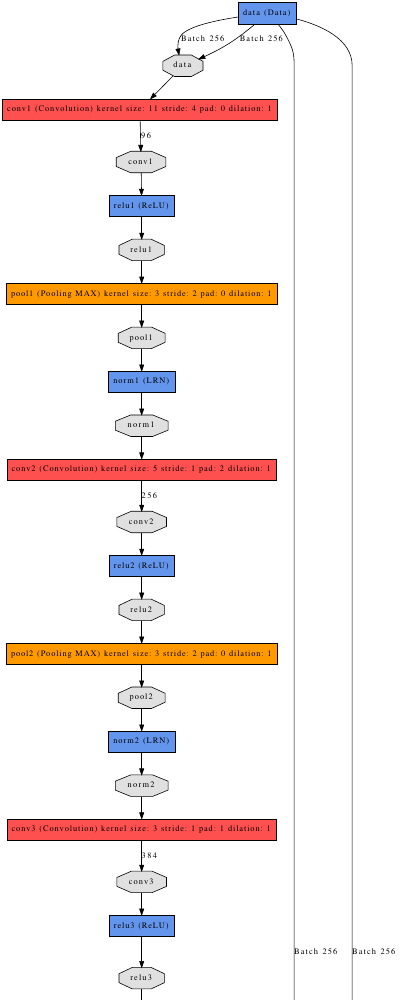}
		\label{fig:imagenet1}
	\end{subfigure}%
	\begin{subfigure}{.5\textwidth}
		\centering
		\includegraphics[scale=1.7]{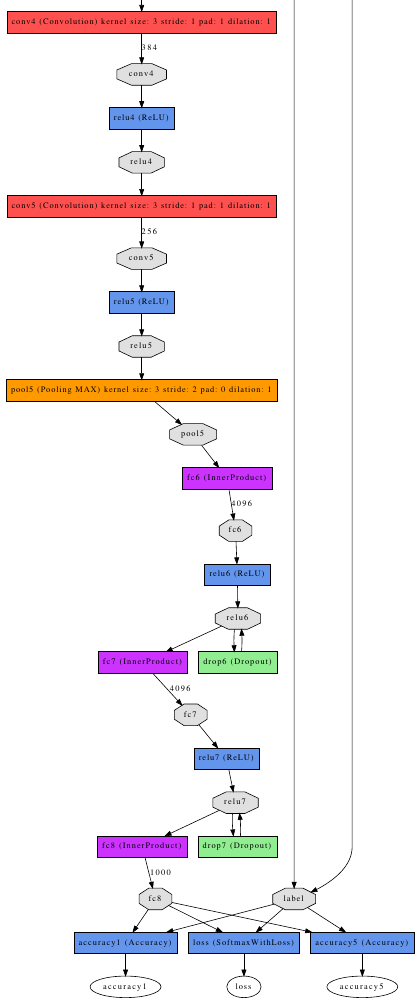}
		\label{fig:imagenet2}
	\end{subfigure}
	\caption{Standard AlexNet/CaffeNet/ImageNet \cite{AlexNet2012}}
\end{figure}
\newpage

\section{Caffe-ImageNet-MOE}
\label{sec:app_imagenet_moe}
\subsection{Main Network}
\begin{figure}[H]
	\centering
	\includegraphics[scale=1.0]{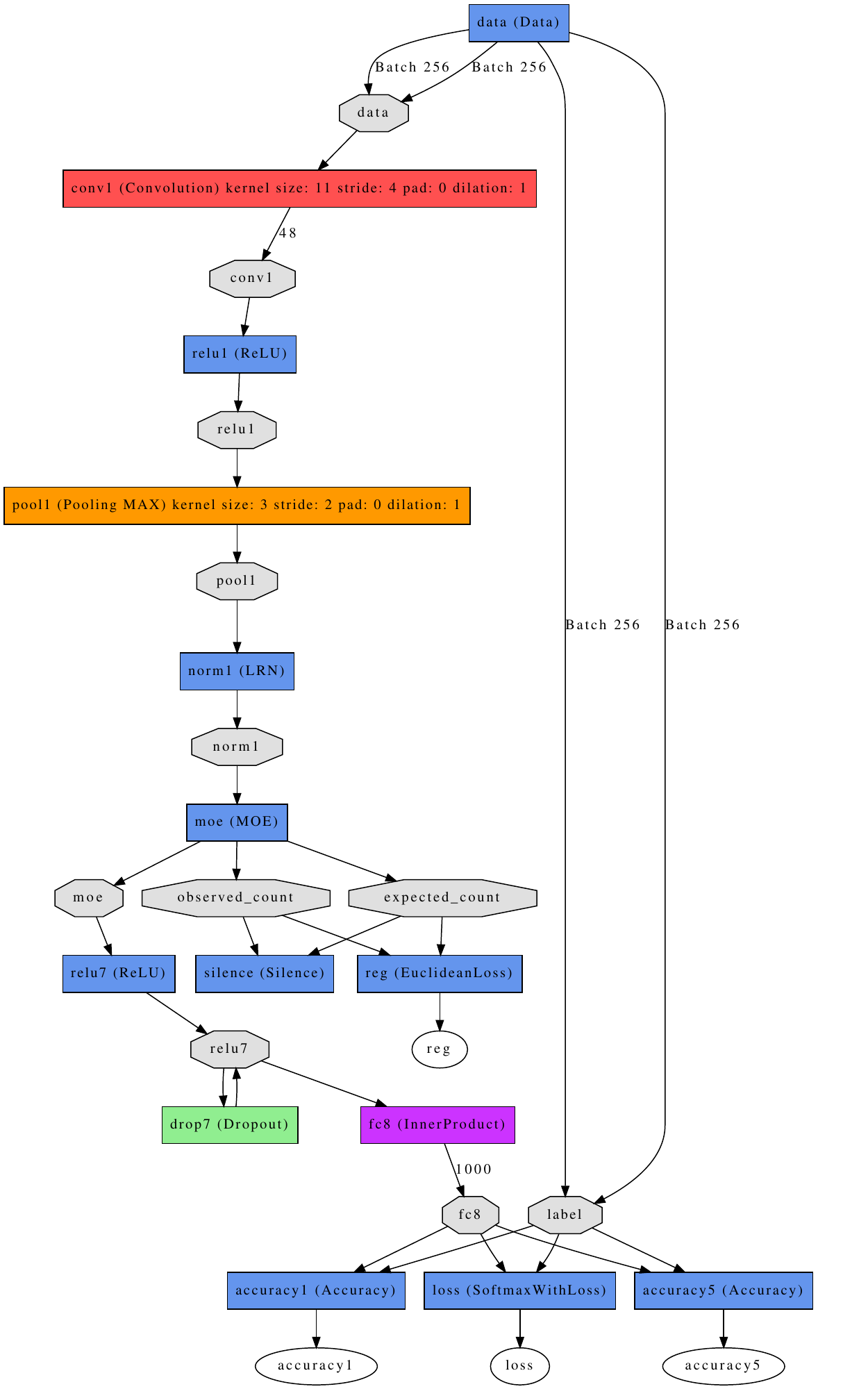}\\
	\label{fig:imagenet_moe_1}
	\caption{Main mixture-of-experts ImageNet}
\end{figure}
\newpage

\subsection{Gating Network}
The gating network is executed within the MOE-Layer.
\begin{figure}[H]
	\centering
	\includegraphics[scale=1.6]{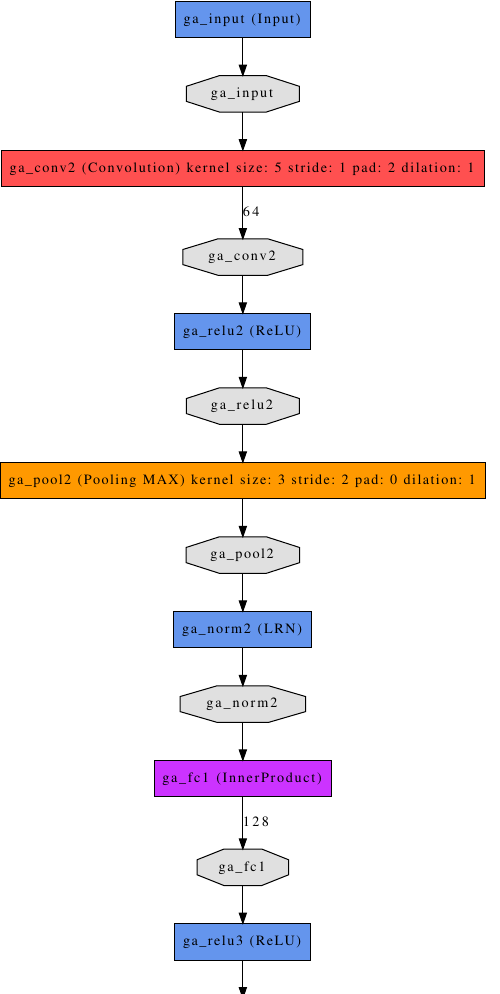}
	\label{fig:imagenet_moe_4}
	\caption{Gating network of the mixture-of-experts ImageNet}
\end{figure}
\begin{figure}[H]
	\centering
	\includegraphics[scale=1.4]{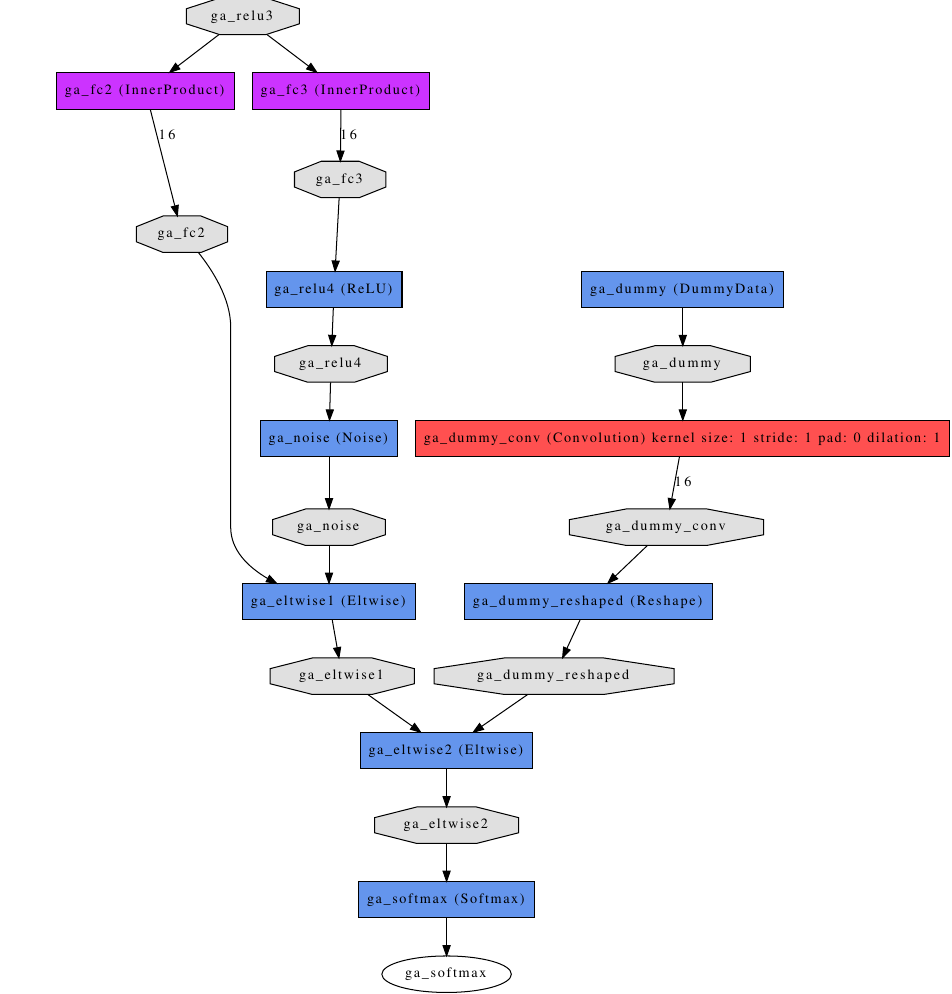}
	\label{fig:imagenet_moe_5}
	\caption{Gating network of the mixture-of-experts ImageNet}
\end{figure}
\newpage

\subsection{Expert Network}
The expert network is repeated 16 times within the MOE-Layer.
\begin{figure}[H]
	\centering
	\begin{subfigure}{.5\textwidth}
		\centering
		\includegraphics[scale=1.6]{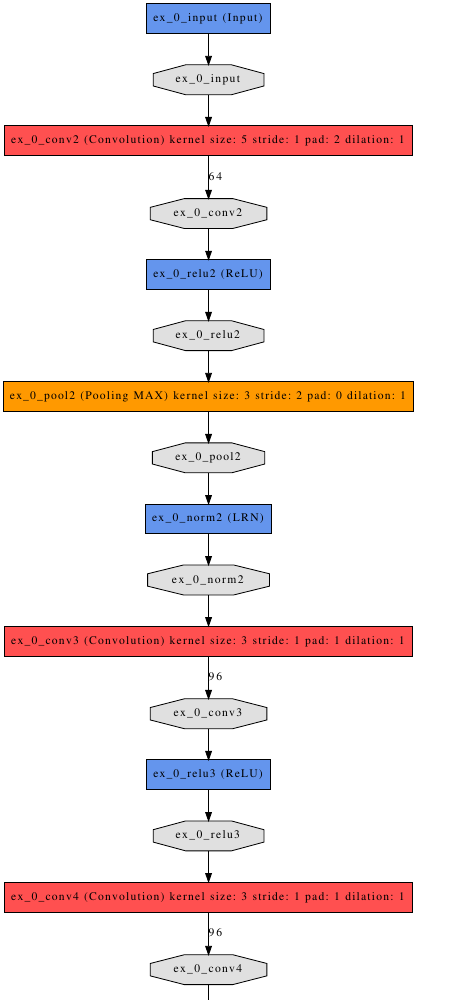}
		\label{fig:imagenet_moe_2}
	\end{subfigure}%
	\begin{subfigure}{.5\textwidth}
		\centering
		\includegraphics[scale=1.6]{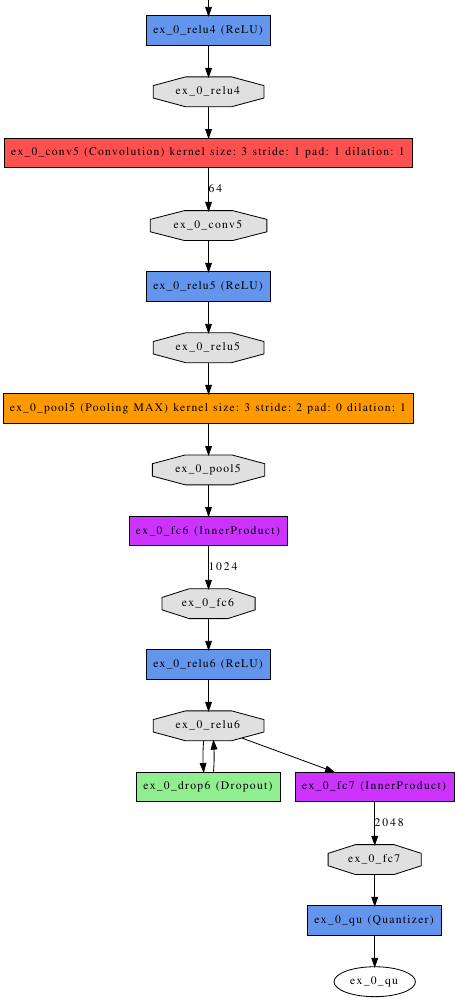}
		\label{fig:imagenet_moe_3}
	\end{subfigure}
	\caption{Expert network of the mixture-of-experts ImageNet}
\end{figure}

\backmatter

\chapter*{References}
\nocite{*}
\printbibliography[heading=none]

@article{Bengio2015,
  author = {Bengio, Emmanuel and Bacon, Pierre-Luc and Pineau, Joelle and Precup, Doina},
  year = {2015},
  month = {11},
  title = {Conditional Computation in Neural Networks for faster models}
}

@online{armcompute,
  title = {ARM Compute Library},
  url = {https://developer.arm.com/technologies/compute-library},
  urldate = {2018-06-10}
}

@online{raspberrypi,
  title = {Raspberry Pi},
  url = {https://www.raspberrypi.org/},
  urldate = {2018-06-10}
}

@online{asustinkerboard,
  title = {Asus Tinkerboard},
  url = {https://www.asus.com/Single-Board-Computer/Tinker-Board/},
  urldate = {2018-06-10}
}

@online{VC4C,
  title = {Raspberry Pi VideoCore IV OpenCL},
  url = {https://github.com/doe300/VC4CL},
  urldate = {2018-06-10}
}

@inproceedings{Jacob_2018_CVPR,
author = {Jacob, Benoit and Kligys, Skirmantas and Chen, Bo and Zhu, Menglong and Tang, Matthew and Howard, Andrew and Adam, Hartwig and Kalenichenko, Dmitry},
title = {Quantization and Training of Neural Networks for Efficient Integer-Arithmetic-Only Inference},
booktitle = {The IEEE Conference on Computer Vision and Pattern Recognition (CVPR)},
month = {June},
year = {2018}
}

@article{jia2014caffe,
  Author = {Jia, Yangqing and Shelhamer, Evan and Donahue, Jeff and Karayev, Sergey and Long, Jonathan and Girshick, Ross and Guadarrama, Sergio and Darrell, Trevor},
  Journal = {arXiv preprint arXiv:1408.5093},
  Title = {Caffe: Convolutional Architecture for Fast Feature Embedding},
  Year = {2014}
}

@online{BVLCCaffe,
  title = {BVLC Caffe},
  url = {https://github.com/BVLC/caffe},
  urldate = {2018-06-10}
}

@online{Caffe,
  title = {Caffe Improved},
  author = {Fabian Tschopp},
  url = {https://github.com/naibaf7/caffe},
  urldate = {2018-06-10}
}

@online{CaffeExamples,
  title = {Caffe Examples},
  author = {Fabian Tschopp},
  url = {https://github.com/naibaf7/caffe_neural_tool},
  urldate = {2018-06-10}
}

@online{AMD,
  title = {AMD},
  url = {https://www.amd.com/},
  urldate = {2018-06-10}
}

@online{amdhip,
  title= {AMD HIP},
  url = {https://gpuopen.com/compute-product/hip-convert-cuda-to-portable-c-code/},
  urldate = {2018-06-10}
}

@online{Intel,
  title = {Intel},
  url = {https://www.intel.com/},
  urldate = {2018-06-10}
}

@online{GTX1080Whitepaper,
  title = {nVidia GTX 1080 Whitepaper},
  url = {https://international.download.nvidia.com/geforce-com/international/pdfs/GeForce_GTX_1080_Whitepaper_FINAL.pdf},
  urldate = {2018-06-10}
}

@online{VegaWhitepaper,
  title = {AMD Vega Whitepaper},
  url = {https://radeon.com/_downloads/vega-whitepaper-11.6.17.pdf},
  urldate = {2018-06-10}
}

@online{PolarisWhitepaper,
  title = {AMD Polaris Whitepaper},
  url = {http://radeon.com/_downloads/polaris-whitepaper-4.8.16.pdf},
  urldate = {2018-06-10}
}

@online{ARMMaliGPUOpenCL,
  title = {ARM Mali GPU OpenCL},
  url = {http://infocenter.arm.com/help/topic/com.arm.doc.100614_0300_00_en/arm_mali_gpu_opencl_developer_guide_100614_0300_00_en.pdf},
  urldate = {2018-06-10}
}

@online{IntelIrisProGraphics,
  title = {Intel Iris Pro OpenCL},
  url = {https://www.khronos.org/assets/uploads/developers/library/2013-siggraph-opencl-bof/OpenCL-Intel-BOF_SIGGRAPH-2013.pdf},
  urldate = {2018-06-10}
}

@online{GongzgFP16,
  title = {Caffe FP16 Contribution},
  url = {https://github.com/BVLC/caffe/pull/5745},
  urldate = {2018-06-10}
}

@online{cudadp4dp2,
  title = {Mixed-Precision CUDA},
  url = {https://devblogs.nvidia.com/mixed-precision-programming-cuda-8/},
  urldate = {2018-06-10}
}

@article{Tschopp2015,
  author    = {Fabian Tschopp},
  title     = {Efficient Convolutional Neural Networks for Pixelwise Classification
               on Heterogeneous Hardware Systems},
  journal   = {CoRR},
  volume    = {abs/1509.03371},
  year      = {2015},
  url       = {http://arxiv.org/abs/1509.03371},
  archivePrefix = {arXiv},
  eprint    = {1509.03371},
  bibsource = {dblp computer science bibliography, https://dblp.org}
}

@inproceedings{Tschopp2016,
  doi = {10.1109/isbi.2016.7493487},
  url = {https://doi.org/10.1109/isbi.2016.7493487},
  year  = {2016},
  month = {apr},
  publisher = {{IEEE}},
  author = {Fabian Tschopp and Julien N. P. Martel and Srinivas C. Turaga and Matthew Cook and Jan Funke},
  title = {Efficient convolutional neural networks for pixelwise classification on heterogeneous hardware systems},
  booktitle = {2016 {IEEE} 13th International Symposium on Biomedical Imaging ({ISBI})}
}

@inproceedings{Teera2016,
  doi = {10.1109/icpr.2016.7900006},
  url = {https://doi.org/10.1109/icpr.2016.7900006},
  year  = {2016},
  month = {dec},
  publisher = {{IEEE}},
  author = {Surat Teerapittayanon and Bradley McDanel and H.T. Kung},
  title = {{BranchyNet}: Fast inference via early exiting from deep neural networks},
  booktitle = {2016 23rd International Conference on Pattern Recognition ({ICPR})}
}

@ARTICLE{CuDNN2014,
   author = {{Chetlur}, S. and {Woolley}, C. and {Vandermersch}, P. and {Cohen}, J. and 
	{Tran}, J. and {Catanzaro}, B. and {Shelhamer}, E.},
    title = "{cuDNN: Efficient Primitives for Deep Learning}",
  journal = {ArXiv e-prints},
archivePrefix = "arXiv",
   eprint = {1410.0759},
     year = 2014,
    month = oct,
   adsurl = {http://adsabs.harvard.edu/abs/2014arXiv1410.0759C}
}

@incollection{AlexNet2012,
	title = {ImageNet Classification with Deep Convolutional Neural Networks},
	author = {Alex Krizhevsky and Sutskever, Ilya and Geoffrey E. Hinton},
	booktitle = {Advances in Neural Information Processing Systems 25},
	editor = {F. Pereira and C.J.C. Burges and L. Bottou and K.Q. Weinberger},
	pages = {1097--1105},
	year = {2012},
	publisher = {Curran Associates, Inc.},
	url = {http://papers.nips.cc/paper/4824-imagenet-classification-with-deep-convolutional-neural-networks.pdf}
}

@inproceedings{RuppViennaCL,
	author = {Rupp, K. and Rudolf, F. and Weinbub, J.},
	title = {{ViennaCL - A High Level Linear Algebra Library for GPUs and Multi-Core CPUs}},
	booktitle = {Intl.~Workshop on GPUs and Scientific Applications},
	year = {2010},
	pages = {51-56}
}

@ARTICLE{MobileNets2017,
   author = {{Howard}, A.~G. and {Zhu}, M. and {Chen}, B. and {Kalenichenko}, D. and 
	{Wang}, W. and {Weyand}, T. and {Andreetto}, M. and {Adam}, H.
	},
    title = "{MobileNets: Efficient Convolutional Neural Networks for Mobile Vision Applications}",
  journal = {ArXiv e-prints},
archivePrefix = "arXiv",
   eprint = {1704.04861},
 primaryClass = "cs.CV",
     year = 2017,
    month = apr,
   adsurl = {http://adsabs.harvard.edu/abs/2017arXiv170404861H}
}

@inproceedings{Benitez1997,
  author = {J.  M.  Benitez,  J.  L.  Castro,  and  I.  Requena},
  title = {Are Artificial Neural Networks Black Boxes?},
  journal = {IEEE  TRANSACTIONS  ON  NEURAL  NETWORKS,  VOL.  8,  NO.  5,  SEPTEMBER  1997},
  year = 1997,
  month = sep
}

@article{lecun-mnisthandwrittendigit-2010,
  author = {LeCun, Yann and Cortes, Corinna},
  howpublished = {http://yann.lecun.com/exdb/mnist/},
  lastchecked = {2016-01-14 14:24:11},
  title = {{MNIST} handwritten digit database},
  url = {http://yann.lecun.com/exdb/mnist/},
  username = {mhwombat},
  year = 2010
}

@inproceedings{LeNet1998,
	author = {Y. LeCun, L. Bottou, Y. Bengio, and P. Haffner. },
	title = {{Gradient-based learning applied to document recognition.}},
	booktitle = {Proceedings of the IEEE},
	year = {1998},
}

@online{intelark,
	title={Intel ARK database},
	url = {http://ark.intel.com/},
	urldate = {2018-06-10}
}

@inproceedings{Nugteren2018,
  doi = {10.1145/3204919.3204924},
  url = {https://doi.org/10.1145/3204919.3204924},
  year  = {2018},
  publisher = {{ACM} Press},
  author = {Cedric Nugteren},
  title = {{CLBlast}},
  booktitle = {Proceedings of the International Workshop on {OpenCL}  - {IWOCL} {\textquotesingle}18}
}

@inproceedings{Shazeer2017,
	title = {Outrageously Large Neural Networks: The Sparsely-Gated Mixture-of-Experts Layer},
	author  = {Noam Shazeer and Azalia Mirhoseini and Krzysztof Maziarz and Andy Davis and Quoc Le and Geoffrey Hinton and Jeff Dean},
	year  = {2017},
  booktitle = {ICLR},
	URL = {https://openreview.net/pdf?id=B1ckMDqlg}
}

\end{document}